%% file: cpp.tex
\documentclass[10pt,twocolumn,letterpaper]{article}

\usepackage{iccv}
\usepackage{times}
\usepackage{epsfig}
\usepackage{graphicx}
\usepackage{amsmath}
\usepackage{amssymb}

\usepackage[breaklinks=true,colorlinks,bookmarks=false]{hyperref}

\input{math_commands.tex}

\usepackage{hyperref}
\usepackage{url}
\usepackage{pifont}
\usepackage{bbm}
\usepackage{microtype}
\usepackage{booktabs} 
\usepackage{soul}
\usepackage{xspace}
\usepackage{multirow}
\usepackage{graphbox}
\usepackage{array}
\usepackage[capitalize,noabbrev]{cleveref}
\usepackage[ruled,vlined]{algorithm2e}
\usepackage{cuted}

\newcommand{\ours}{CPP\xspace}
\newcommand{\ourloss}{CPL\xspace}
\newcommand{\xmark}{\ding{55}}%
\newcommand{\authorskip}{\qquad}

\usepackage[breaklinks=true,bookmarks=false]{hyperref}

\iccvfinalcopy 

\def\iccvPaperID{****} 

\begin{document}

\title{Steering Prototypes with Prompt-tuning for Rehearsal-free Continual Learning}

\author{Zhuowei Li\textsuperscript{1} \authorskip Long Zhao\textsuperscript{2} \authorskip Zizhao Zhang\textsuperscript{3} \authorskip Han Zhang\textsuperscript{4} \authorskip Di Liu\textsuperscript{1}\\ Ting Liu\textsuperscript{2} \authorskip Dimitris N.\ Metaxas\textsuperscript{1}\\[2mm]
\textsuperscript{1}Rutgers University \authorskip \textsuperscript{2}Google Research \authorskip \textsuperscript{3}Google Cloud AI \authorskip \textsuperscript{4}{Google DeepMind}}

\maketitle

\begin{abstract}
In the context of continual learning, prototypes—as representative class embeddings—offer advantages in memory conservation and the mitigation of catastrophic forgetting. However, challenges related to semantic drift and prototype interference persist. In this study, we introduce the \textbf{C}ontrastive \textbf{P}rototypical \textbf{P}rompt (CPP) approach. Through task-specific prompt-tuning, underpinned by a contrastive learning objective, we effectively address both aforementioned challenges. Our evaluations on four challenging class-incremental benchmarks reveal that CPP achieves a significant 4\% to 6\% improvement over state-of-the-art methods. Importantly, CPP operates without a rehearsal buffer and narrows the performance divergence between continual and offline joint-learning, suggesting an innovative scheme for Transformer-based continual learning systems \footnote{Code is available at \url{https://github.com/LzVv123456/Contrastive-Prototypical-Prompt}.}.
\end{abstract}

\section{Introduction}
Continual learning~\cite{Thrun95k}, defined as the model's ability to sequentially assimilate information from a continuous stream of potentially correlated data, is critical for modern intelligent systems, due to the inherently dynamic nature of the real world~\cite{EmbracingChange}. Yet, existing deep neural networks are known to be prone to \textit{catastrophic forgetting}~\cite{CF1}, \ie models progressively degrade in performance on previously mastered tasks upon the acquisition of new information.

\begin{figure}[t]
\begin{center}
\includegraphics[width=0.95\linewidth]{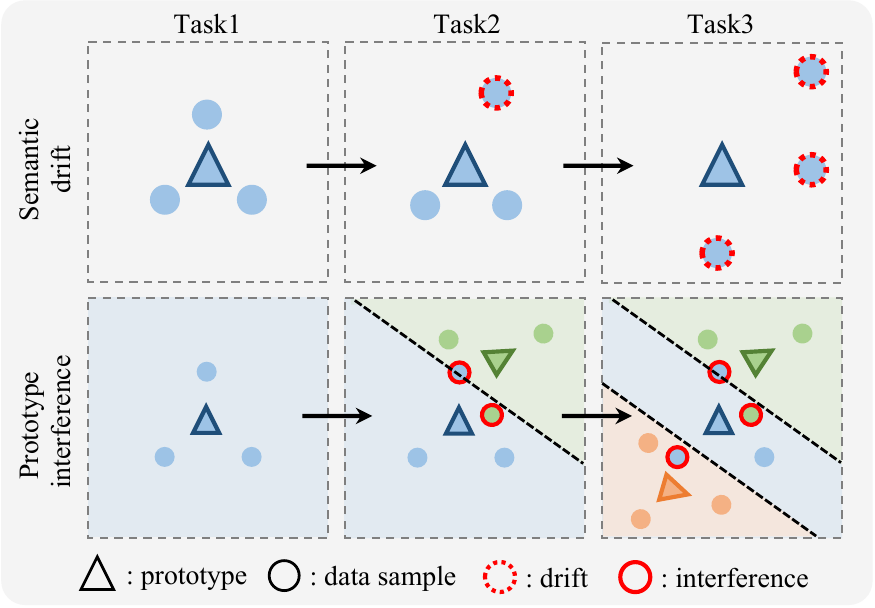}
\end{center}
\caption{A schematic illustration of \textit{semantic drift} and \textit{prototype interference}. Different colors represent separate classes.}
\vspace{-10pt}
\label{fig:proto_forget}
\end{figure}

Among the plethora of strategies proposed to harness continual learning challenges~\cite{3cl}, prototypes (\eg, class mean embeddings~\cite{proto_net}) exhibit a promising functionality as they can retain previous knowledge in a memory-efficient manner~\cite{PROTO_AUG} and mitigate bias towards the latest task~\cite{iCaRL, Yu2020SemanticDC} when coupled with a nearest class mean (NCM)~\cite{NCM} classifier. However, prototypes themselves are also subject to abrupt efficacy drops due to the following challenges:

\begin{itemize}
\item \textbf{Semantic drift}. 
Sequential task learning using a unified model essentially results in a series of model snapshots, with only the most recent version being retained. This framework introduces a discrepancy between the data sample's embedding at inference time and its prototype—leading to pronounced semantic drift in the embedding space (see Fig.~\ref{fig:proto_forget} top).

\item \textbf{Prototype interference}. The emergence of new data samples, bearing high semantic resemblance to previous samples, can distort their embeddings in proximity to pre-existing prototypes, thereby causing interference (see Fig.~\ref{fig:proto_forget} bottom).

\end{itemize}
During course of continual learning, both phenomena occur simultaneously and cause \textit{forgetting}.

Although existing methodologies attempt to alleviate the impact of semantic drift—either through the preservation and replay of prior samples~\cite{iCaRL, De_Lange_2021_ICCV} or by compensating for drifts post-facto~\cite{Yu2020SemanticDC,Gu_Deng_Wei_2021}—they often necessitate significant memory overhead or become susceptible to cumulative errors as the sequence of tasks extends. Furthermore, the interference issue is less attended to, and is often left implicitly to additional regularization terms~\cite{MAS, LwF, EMC}. To bridge these gaps, our approach seeks to synchronously align the embedding functions employed for prototype generation and inference, thereby avoiding semantic drift. Concurrently, we also advocate for a strategic regulation of distances between prototypes in the embedding space to minimize interference.

Drawing inspiration from prompt-tuning~\cite{prompt_tuning,VPT}, a new transfer learning paradigm that enables a tiny portion of additional learnable tokens to adapt a frozen Transformer~\cite{transformer} to down-stream tasks, we propose the use of task-specific prompts to bypass semantic drift. Concretely, we allow a data sample at inference to retrieve its corresponding task-specific prompts and assemble the exact embedding function that is used for generating its corresponding prototype. Under our schema, a frozen, pre-trained Transformer is treated as shared, stabilized global knowledge across tasks, ensuring system stability. Task-specific prompts, on the other hand, learn task-level specializations and keep the system plastic.

To reduce prototype interference in the embedding space, we propose a novel contrastive prototypical loss optimized using task-specific prompts. The crafted learning objective encourages intra-class clustering and increases inter-class distances upon on a mixture of data embeddings and up-to-now prototypes. Our architecture ensures that historical knowledge is assimilated and utilized solely as prototypes and anchor points. Task-specific prompts can, therefore, effectively navigate current prototypes to minimize interference, obviating the need for the storage and replay of explicit samples. We further enhance our framework with the multi-centroid prototype strategy, which deploys a set of fictive embeddings instead of a singular mean embedding, capturing the distributional essence of a class more comprehensively.

We term our method \textbf{C}ontrastive \textbf{P}rototypical \textbf{P}rompt (\ours), a simple and novel continual learning framework that explores embedding space holistically. Emperically, \ours excels in four challenging class-incremental benchmarks including split CIFAR-100, 5-datasets, split ImageNet-subset, and split ImageNet-R, bringing around 4\% to 6\% absolute improvements over state-of-the-art methods. Notably, \ours is rehearsal-free and consumes at least $5\times$ times fewer additional memories than rivals. Our primary achievements can be summarized as:
\begin{itemize}
	\item The inception of \ours, a simple and novel framework for rehearsal-free continual learning. It leverages contrastively learned task-specific prompts to effectively address both semantic drift and prototype interference obstacles.
	\item The introduction of the multi-centroid prototype strategy, an innovative mechanism that enriches the representational prowess of prototypes and is seamlessly integrated into the \ours framework.
	\item Rigorous empirical validations confirming the supremacy of \ours under a light memory budget. Each crafted component is thoroughly studied and demonstrates clear and additive benefits.
\end{itemize}

\section{Related Work}
\noindent\textbf{Continual learning}. Prevalent algorithms can typically be categorized into three primary branches~\cite{PARISI201954, 3cl, EmbracingChange}. Regularization-based methods strike a balance under \textit{stability–plasticity dilemma}. A subset of this, parameter-regularization methods, either restrict the magnitude~\cite{EMC, SI} or the orientation~\cite{OGD, GPM} of parameter space alterations. Conversely, functional-regularization methods confine functional changes concerning certain 'anchor points' in the function space~\cite{FR, titsias2019functional, NEURIPS2020_2f3bbb97}. Notably, their performance tends to wane with extended task sequences~\cite{EmbracingChange}. Modular-based methods mitigate knowledge interference by either reallocating existing resources~\cite{PackNet, pmlr-v80-serra18a} or provisioning additional learning capacities~\cite{PNN, yoon2018lifelong, Li2019LearnTG}. A caveat is their frequent dependence on test-time task identifiers, making them challenging to scale. In practice, by saving and replaying earlier data instances, rehearsal-based methods have consistently demonstrated versatility and resilience~\cite{dark_memory, Co2L}. However, their efficacy can vary based on buffer size~\cite{GDumb, EmbracingChange} and might not be practical for scenarios with tight memory constraints or heightened privacy concerns. Our proposed \ours, integrates the advantages of these diverse methods while strategically sidestepping their inherent limitations.

\noindent\textbf{Prototypes for continual learning}. Embeddings have been recognized for their resilience against information loss~\cite{Yu2020SemanticDC, Davari_2022_CVPR}. Moreover, the parameter-driven linear classifiers often culminate in rapid forgetting, chiefly due to biases in favor of recent tasks~\cite{Zhang_2021_CVPR}. Consequently, a majority of prototype-centric methodologies have harnessed prototypes in combination with the NCM classifier~\cite{iCaRL,Yu2020SemanticDC, PROTO_AUG}. An alternative strategy, PASS~\cite{PROTO_AUG}, employs prototypes as latent space anchors, curtailing semantic overlaps. Similarly, CPP also deploys prototypes as anchors in the latent space. To empower the prototypes, existing approaches to counteract semantic drift have either preserved explicit exemplars for prototype updates~\cite{iCaRL, De_Lange_2021_ICCV} or retroactively compensated for drifts by inferring them from contemporary data~\cite{Yu2020SemanticDC}. In stark contrast, \ours preemptively obstructs semantic drifts and proactively tackles prototype interference. Further, our methodology favors multi-centroid prototypes over mean embeddings, offering a nuanced representation of contextual distributions.

\noindent\textbf{Prompt tuning}. The practice of initializing deep neural networks with pre-trained weights is widely adopt. However, conventional fine-tuning might not consistently enhance performance on downstream tasks~\cite{kumar2022finetuning}. In response, prompt-tuning, initially gaining traction in the NLP domain~\cite{prefix_tuning, prompt_tuning} followed by the adaption to the visual domain~\cite{VPT}, has been explored. Its recent induction into continual learning has been evidenced by methodologies such as L2P~\cite{l2p} and DualPromt~\cite{DUAL_PROMPT}, which capitalize on shared prompt pools or universal prompts for incremental knowledge acquisition. In a similar vein, S-prompts~\cite{S-prompt} harnesses domain-specific prompts for domain-incremental learning challenges. Our approach uses task-specific prompts to address semantic drift and prototype interference. Importantly, we innovatively integrate prompt-tuning with prototypes, optimizing their combined benefits for continual learning.

\section{Methodology}
\begin{figure*}[t]
	\begin{center}
		\includegraphics[width=0.9\textwidth]{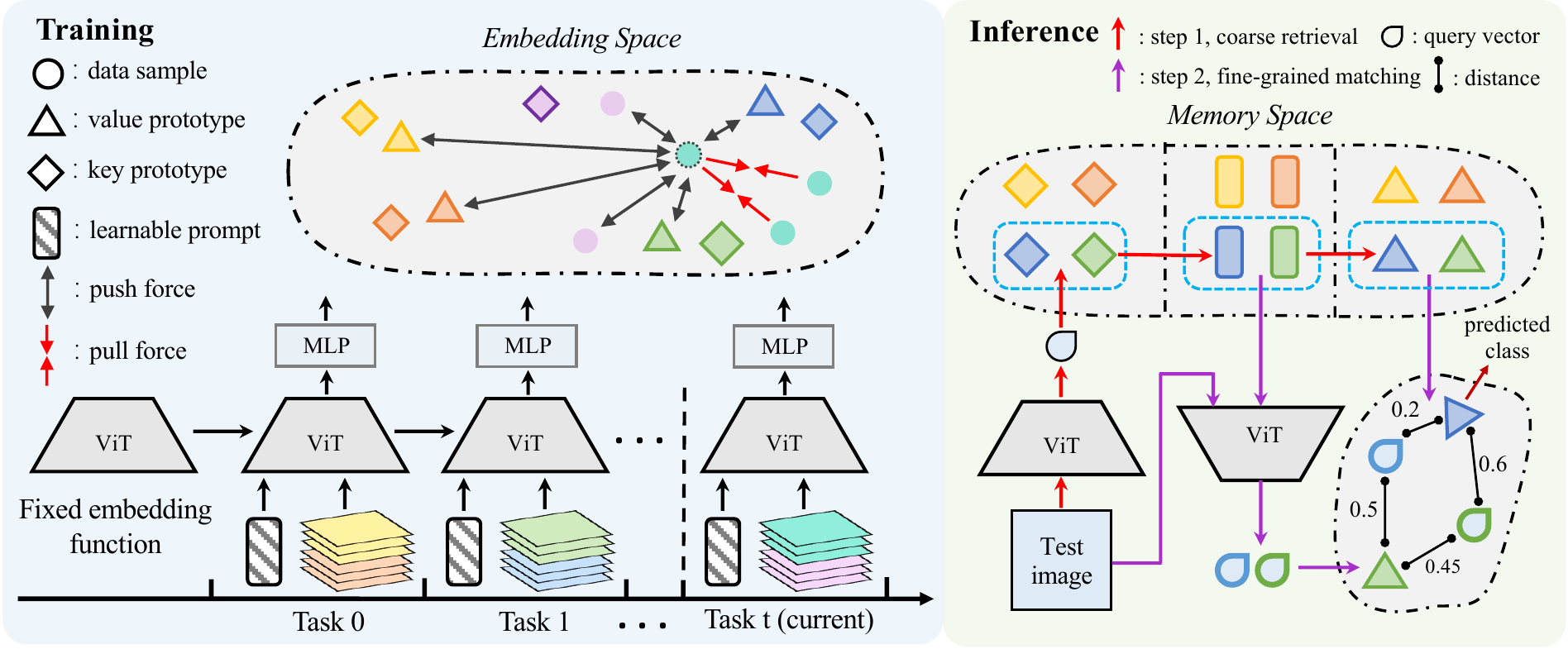}
	\end{center}
	\caption{An overview of \ours. Different colors denote different classes. \textbf{Left:} As learning progresses, knowledge from prior tasks is anchored as prototypes in the embedding space, with task-specific prompts minimizing prototype interference. \textbf{Right:} During inference, a group of candidate prompts (shown as class-specific prompts for illustration purpose) are first retrieved and then followed by a fine-grained matching process.}
	\vspace{-10pt}
	\label{fig:overview}
\end{figure*}

We begin by describing the problem setup and, along the way, introduce the notations in Sec.~\ref{sec:setup}. We then proceed to a basic prototype-based framework in Sec.\ref{sec:baseline}, serving as our baseline model. Building on this baseline, Sec.\ref{sec:cpp} describes our advanced \ours approach. The multi-centroid prototype strategy is further discussed in Sec.\ref{sec:multi-centroid}. An illustrative framework overview can be seen in Fig.~\ref{fig:overview}.

\subsection{Problem Setup and Notion}\label{sec:setup}
Supervised continual learning can be defined as learning a model over a sequence of $T$ tasks $\mathcal{T}_{1:T}=\{\mathcal{T}_1, \mathcal{T}_2, \dots, \mathcal{T}_T\}$. Each task $\mathcal{T}_t$ is associated with a dataset $\mathcal{D}^{t}=\{(\vx_{i}^{t}, y_{i}^{t})_{i=1}^{n_{t}}\}$ containing $n_{t}$ data pairs, where $\vx$ is the input and $y$ is its corresponding label. Each data pair $(\vx_{i}^{t}, y_{i}^{t}) \in  (\mathcal{X}^{t}\times\mathcal{Y}^{t})$ belongs to an unknown distribution $(\mathcal{X}^{t}\times\mathcal{Y}^{t})$ and $\mathcal{Y}^{t} \cap \mathcal{Y}^{t'}=\emptyset$ while ${t}\neq{t'}$. In general, a neural network at session $t$ can be split into an embedding function $f_{\theta^t}(\cdot):\R^{W\times H\times C}\rightarrow \R^D$ and a classifier $g_{\phi^t}(\cdot):\R^D\rightarrow \R^K$ that parameterized by $\theta^t$ and $\phi^t$, respectively. The overall learning objective is to develop a pair of $f_{\theta^t}(\cdot)$ and $g_{\phi^t}(\cdot)$ that excels in all prior tasks $\mathcal{T}_{1:t}$. Note that in this paper, we focus on the challenging rehearsal-free class-incremental setting, where the task identity $t$ is not available during inference and we do not augment the dataset $\mathcal{D}^t$ of task $\mathcal{T}_t$ by saving and replaying preceding exemplars during the training stage. 

\subsection{A Training-free Baseline}\label{sec:baseline}
Let $\mathcal{D}_k^t$ denote the dataset belonging to class $k$ at task $t$, and the prototype of class $k$ is produced as the class mean embedding following \cite{iCaRL}:
\begin{equation}
	\label{eq:class_mean}
	\vmu_k = \frac{1}{|\mathcal{D}_k^t|}\sum_{\vx \in \mathcal{D}_k^t} f_{\theta}(\vx).
\end{equation}
Here, $\theta$ is initialized by a pre-trained ViT~\cite{vit} and is kept frozen throughout the entire learning process. We then do classification using the nearest-class-mean (NCM)~\cite{NCM} classifier:
\begin{equation}
	\label{eq:argmax}
	y^*=\underset{y\in\{1, \dots, K\}}{\argmin}\{d(\vu_y, f_\theta(\vx))\},
\end{equation} 
where $d:\R^{D} \times \R^{D} \rightarrow \R$ is a distance measurement. In our design, we always define $d$ as the cosine distance (similarity). This straightfoward training-free baseline model produces decent results under an apt embedding function (see Table~\ref{table:main_ablation}), highlighting the value of the embedding and the potential of prototypes in continual learning.

\subsection{Contrastive Prototypical Prompt}\label{sec:cpp}

\subsubsection{Steering prototypes with task-specific prompts} 
Ideally, a static embedding function can position data samples close to their corresponding prototypes in the embedding space, thus avoiding forgetting with the architecture outlined in Sec.~\ref{sec:baseline}. Yet, achieving a universally adept embedding function remains elusive. Instead, we permit the embedding function to adapt dynamically to new inputs with minimal forgetting. This is achieved by incorporating a tiny set of learnable tokens (\ie, task-specific prompts), to inform a frozen embedding function.

Specifically, we prepend a prompt $\vp_i \in \R^{L_p\times D}$ to the existing tokens in the $i$-th layer of a Transformer, where $L_p$ is the length of the prompt and $D$ denotes the embedding dimension. The computation of the $i$-th layer is then defined as:
\begin{equation}
	[\vc_i, \ve_i] = T_i([\vc_{i-1}, \vp_{i-1}, \ve_{i-1}]),
\end{equation}  
where $T_i$ represents a multi-head self-attention block followed by a feed-forward network. Here, $\vc\in\R^{1\times D}$ denotes the class token, and $\ve\in\R^{L_e\times D}$ is the existing data tokens. The operator $[\cdot]$ concatenates along the sequence length dimension. We adopt deep prompt~\cite{VPT} by adding prompts to all $S$ layers. Thus the task-specific prompt for task $t$ is given as $P^t=\{\vp^t_1, \vp^t_2, \dots, \vp^t_S\}$, and the embedding function for task $t$ can be rewritten as:
\begin{equation}
	f_{\theta^t}(\cdot) \rightarrow f_{\{\theta, P^t\}}(\cdot).
\end{equation}
We maintain a collection of task-specific prompts in the memory space, each paired with distinct key and value prototypes (elaborated in Sec.~\ref{sec:inference}). Thanks to the design of task-specific prompts, each task has a unique parameter space and the semantic drifts can be avoided by assembling the frozen embedding function with the correct task-specific prompt at inference.

\subsubsection{Contrastive prototypical loss (\ourloss)}\label{sec:cpl}
To address prototype interference, we introduce a novel loss function tailored to optimize task-specific prompts. Recall that \ours maintains information from different contexts as prototypes and uses the NCM classifier for discrimination. We therefore expect contemporary task-specific prompt can encouraging intra-class clustering while avoiding overlap with prior prototypes in the embedding space.

Based on this rationale, we present \ourloss which fortifies intra-class coherence while distancing inter-class entities within a joint space of data embeddings and prototypes. For task $t$, let $I=\{(\vx_1, y_1),\dots,(\vx_N, y_N)\}$ denote a batch of $N$ image pairs and $Z=\{\vz_1, \dots,\vz_N\}$ be their corresponding embeddings. Here, omitting subscripts, $\vz$ is computated as $\vz=m_{\sigma^t}(f_{\{\theta; P^t\}}(\vx))$, where $m_{\sigma^t}(\cdot)$ is a multi-layer perceptron (MLP) neck parameterized by $\sigma^t$. It is worth noting that $m_{\sigma^t}(\cdot)$ is re-initialized for each new task and disposed during inference. The learning objective is defined as:
\begin{gather} 
	\mathcal{L} = \frac{1}{N}\sum_{i\in \{1,\dots, N\}}{\mathcal{L}_i},\\ 
	\mathcal{L}_i = \frac{-1}{|P(i)|}\sum_{\vz_p\in P(i)}\log{\frac{\exp(\vz_i\cdot \vz_p/\tau)}{\underset{\vz_n\in N(i)\cup \hat{U}}\sum\exp(\vz_i \cdot \vz_n/\tau)}},
	\label{eq:proto_con}
\end{gather}
where $P(i)=\{\vz_p \in Z: y_p = y_i\}$ denotes a set of positive samples \wrt embedding $\vz_i$, and $N(i)=\{\vz_n \in Z: y_n \neq y_i\}$ represents a set of negative samples that do not share the same label as $\vz_i$. We define $\hat{U}=\{\hat{\vu}_1,\dots,\hat{\vu}_k\}$ as a collection of negative anchors, represented by value prototypes from preceding classes. Fig.~\ref{fig:overview} (left) illustrates the idea of our learning objective. To better restrain the discrimination boundaries, we apply prototype augmentation following~\cite{PROTO_AUG}. At each iteration, prototypes in $U$ are randomly sampled and perturbed by a scaled Gaussian noise $\hat{\vmu} = \vmu + m * \ve$, where $\ve\sim\mathcal{N}(0, 1)$ and $m$ is a scale factor computed as the average variance of the corresponding class embeddings.

\ourloss differs from the canonical contrastive loss~\cite{SimCLR, SCL} in two primary ways. First, we employ previous class prototypes as negative anchors, preserving spaces for earlier data and negating prototype interference. Second, we concentrate on aligning positive embeddings, omitting the emphasis on intra-class uniformity, a crucial aspect in conventional contrastive learning~\cite{alignment_uniformity}. Given the NCM classifier's inclination to select the nearest prototype, enhancing intra-class uniformity might undesirably widen the distance between a sample and its prototype. This rationale can also be understood from an energy-based standpoint; further details are expounded in the supplementary material.

\subsubsection{Inference by reemerging model snapshots}\label{sec:inference} 
In this section, we delineate the inference procedure using \ours. The crux of the procedure is to associate a target data sample with its task-specific prompt, thereby reconstituting the complete embedding function, i.e., a model snapshot. This process is analog to predicting task identities and can be simplified to directly assigning correct task-specific prompt to a given class when the task identities are available.

Considering the prototype of class $k$ in task $t$, we decouple it into a \textit{key} prototype $\vu_k=\frac{1}{|{D}_k^t|}\sum_{\vx\in\mathcal{D}_k^t}f_{\theta}(\vx)$ and a \textit{value} prototype $\hat{\vu}=\frac{1}{|{D}_k^t|}\sum_{\vx\in\mathcal{D}_k^t}f_{\{\theta, P^t\}}(\vx)$ corresponding to the class mean embedding before and after learning task $t$, respectively. We cache both key prototypes $U = \{\vu_1,\dots,\vu_K\}$ and value prototypes $\hat{U} = \{\hat{\vu}_1,\dots, \hat{\vu}_K\}$ of $K$ learned classes. For inference, we start by constructing a coarse query vector $\vq:\R^{1\times D}$ of target sample $\vx$. We subsequently employ a query function $q(\vq, U, r)$ to locate $r$ nearest key prototypes and retrieve their corresponding prompt sets $\{P^1,\dots,P^J:J\leq r\}$. Here, $J\leq r$ since distinct classes might possess identical task-specific prompts. $\vq$ is simply the class token from the last layer: $\vq = f_{\theta}(\vx)$ (exact indexing operation is omitted to avoid notion clutter) and the query function measures the pairwise cosine similarity between $\vq$ and $U$. 

Deeming these retrieved prompts as candidates, we generate a set of fine-grained query vectors $\hat{Q}=\{\hat{\vq}^j: j\in[1, J]\}$, where $\hat{\vq}^j=f_{\{\theta, P^j\}}(\vx)$. The final class prediction is made as following:
\begin{equation}
	\label{eq:final_predict}
	y^*=\underset{y}{\argmin}\{d(\hat{\vu}_y,\hat{\vq}^j): y\in[1,K], j\in[1,J]\}.
\end{equation}
Fig.~\ref{fig:overview} (right) depicts the information flow of the inference process. Detailed algorithms are provided in supplementary. 

There are two key steps that ensure the success of the above inference process. First, we need to retrieve a group of candidate prompts that contain the target prompt. This is secured by an appropriate embedding function through which a data sample will locate in the vicinity of its corresponding distribution mass center (empirically validated in supplementary). Second, we expect the retrieved prompts, \ie, both target prompt and mismatched prompts, can attribute. Thanks to our designed contrastive prototypical loss, target prompt will reduce the distance between the sample and its corresponding value prototype, while mismatched prompts act in the opposite way. Hence, the behavior of both target and mismatched prompts is beneficial for the NCM.
\begin{figure}[t]
\begin{center}
	\includegraphics[width=0.45\textwidth]{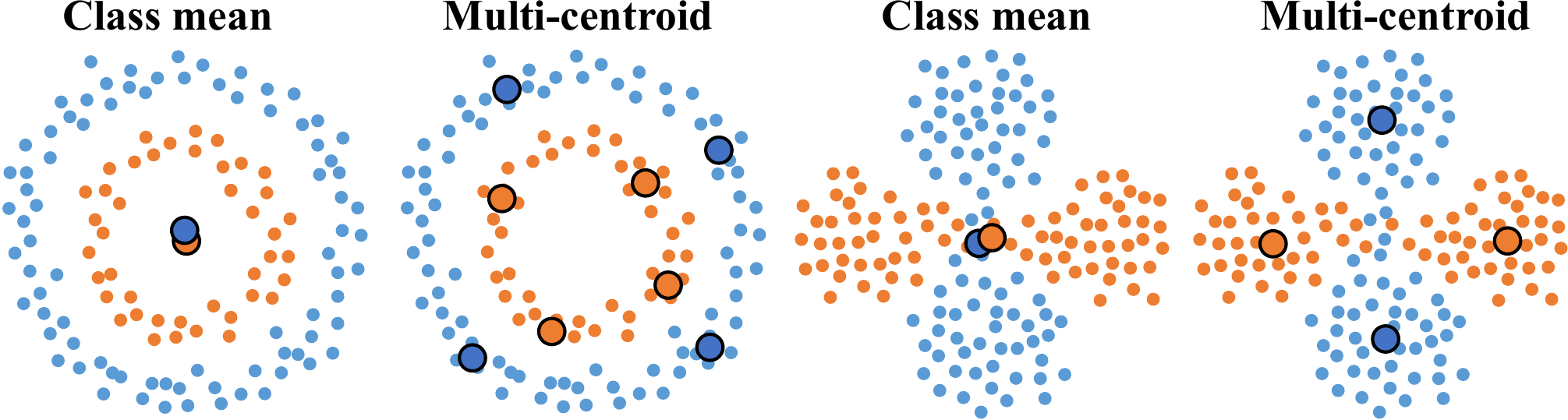}
\end{center}
\caption{Different colors denote different classes. Small and large circles represent data samples and prototypes, respectively.}
\label{fig:multi-centroids}
\end{figure}

\subsection {Multi-centroid Prototypes}\label{sec:multi-centroid}
Prevailing alrothims in continual learning adopt class mean embeddings as prototypes~\cite{Yu2020SemanticDC,PROTO_AUG,FACT}. This practice harbors the belief that latent space distributions are both convex and isotropic. Furthermore, it assumes that the distance function belongs to Bregman divergence~\cite{proto_net}. Yet, above premise don't always stand their ground in practice as no strict constraints are imposed on embedding distributions, and the cosine distance is not Bregman divergence (see Fig.~\ref{fig:multi-centroids} for an example). 

To this end, we propose to use multi-centroid prototype. The essence of this strategy is the generation of a set of synthetic embeddings to encapsulate the quintessence of the target class distribution. Specifically, we calculate the similarity matrix $\mS_k:\R^{|\mathcal{D}_k|\times |\mathcal{D}_k|}$ by measuring the pair-wise cosine similarities between embeddings in class $k$. $\mS_k$ is then used as the affinity matrix for spectral clustering~\cite{spectral_clustering} to generate $C$ centroids $\{\vu_{k,c}\}_{c=1}^C$. In both training and inference stages, we seamlessly substitute class mean embeddings with their corresponding multi-centroid prototypes. 

It is worth noting that the multi-centroid strategy is also complementary to CPP's inference framework serving as a hard-case miner. Consider a datum that either strays far from its distribution's mass center or finds itself mired amidst overlapping class distributions; the multi-centroid approach facilitates the retrieval of a broader spectrum of candidate prompts, thereby safeguarding accuracy. Conversely, for data inhabit near their distribution mass centers in a well-separated class territories, this strategy prunes the list of retrieved prompts, boosting efficiency. The upper and lower bounds of the retrieved prompt number are given as $r$ and $\max(1, r/C)$, respectively.

\begin{table*}[t]
\begin{center}
\resizebox{0.9\linewidth}{!}{%
\begin{tabular}{l @{\hskip +0.1cm}c cc c cc c cc}
	\toprule 
	\multirow{2}{*}{\textbf{Method}} & \multirow{2}{*}{\textbf{Buffer}} & \multicolumn{2}{c}{\textbf{Split CIFAR-100}} &\multirow{2}{*}{\textbf{Buffer}} & \multicolumn{2}{c}{\textbf{5-datasets}} &\multirow{2}{*}{\textbf{Buffer}} & \multicolumn{2}{c}{\textbf{Split ImageNet-R}} \\
	& &  Avg. Acc ($\uparrow$) & Forget ($\downarrow$) & & Avg. Acc ($\uparrow$) & Forget ($\downarrow$) & & Avg. Acc ($\uparrow$) & Forget ($\downarrow$) \\
	\midrule
    ER~\cite{ER} &\multirow{5}{*}{5000 }& 82.53\scriptsize{$\pm$0.17} & 16.46\scriptsize{$\pm$0.25}&\multirow{5}{*}{500}& {84.26\scriptsize{$\pm$0.84}} & 12.85\scriptsize{$\pm$0.62} &\multirow{5}{*}{5000 }& 65.18\scriptsize{$\pm$0.40} & 23.31\scriptsize{$\pm$0.89} \\
	BiC~\cite{BiC} & & 81.42\scriptsize{$\pm$0.85} & 17.31\scriptsize{$\pm$1.02} && 85.53\scriptsize{$\pm$2.06} & 10.27\scriptsize{$\pm$1.32}&& 64.63\scriptsize{$\pm$1.27} & 22.25\scriptsize{$\pm$1.73} \\
	GDumb~\cite{GDumb}  & & 
	81.67\scriptsize{$\pm$0.02} & - &&- &- && 65.90\scriptsize{$\pm$0.28} & -  \\
	DER++~\cite{dark_memory}  & & 83.94\scriptsize{$\pm$0.34} & 14.55\scriptsize{$\pm$0.73} && 84.88\scriptsize{$\pm$0.57} & 10.46\scriptsize{$\pm$1.02} && 66.73\scriptsize{$\pm$0.87} & 20.67\scriptsize{$\pm$1.24} \\
	Co$^2$L~\cite{Co2L}  & & 82.49\scriptsize{$\pm$0.89} & 17.48\scriptsize{$\pm$1.80} && 86.05\scriptsize{$\pm$1.03} & 12.28\scriptsize{$\pm$1.44} && 65.90\scriptsize{$\pm$0.14} & 23.36\scriptsize{$\pm$0.71} \\
	\midrule
	EWC~\cite{EMC} & & 47.01\scriptsize{$\pm$0.29} & 33.27\scriptsize{$\pm$1.17} && 50.93\scriptsize{$\pm$0.09} & 34.94\scriptsize{$\pm$0.07} && 35.00\scriptsize{$\pm$0.43} & 56.16\scriptsize{$\pm$0.88} \\
	LwF ~\cite{LwF} & & 60.69\scriptsize{$\pm$0.63} & 27.77\scriptsize{$\pm$2.17} && 47.91\scriptsize{$\pm$0.33} & 38.01\scriptsize{$\pm$0.28} && 38.54\scriptsize{$\pm$1.23} & 52.37\scriptsize{$\pm$0.64} \\
	L2P~\cite{l2p} & & 83.86\scriptsize{$\pm$0.28} & {7.35\scriptsize{$\pm$0.38}} && 81.14\scriptsize{$\pm$0.93} & {4.64\scriptsize{$\pm$0.52}} &&{61.57\scriptsize{$\pm$0.66}} & {9.73\scriptsize{$\pm$0.47}} \\
	ESN~\cite{ESN} & & 86.34\scriptsize{$\pm$0.52} & 4.76\scriptsize{$\pm$0.14} && 85.71\scriptsize{$\pm$1.47} & 2.58\scriptsize{$\pm$0.61} & & - & - \\
	DualPrompt~\cite{DUAL_PROMPT} & & 86.51\scriptsize{$\pm$0.33} & 5.16\scriptsize{$\pm$0.09} && 88.08\scriptsize{$\pm$0.36} & 2.21\scriptsize{$\pm$0.69} & &  68.13\scriptsize{$\pm$0.49} & 4.68\scriptsize{$\pm$0.20}  \\
	\bf \ours (ours) & & \bf 91.12\scriptsize{$\pm$0.12} & \bf 3.33\scriptsize{$\pm$0.18} && \bf 
	92.92\scriptsize{$\pm$0.17} & \bf 
	0.19\scriptsize{$\pm$0.07} && \bf 74.88\scriptsize{$\pm$0.07} & \bf 3.65\scriptsize{$\pm$0.03} \\
	\midrule
    Upper-bound & -& 93.15\scriptsize{$\pm$0.09} & - & - & 97.81\scriptsize{$\pm$0.02} & - & - & 83.87\scriptsize{$\pm$0.30} & - \\
	\bottomrule
	\end{tabular}
	}
\end{center}
\caption{Comparison to state-of-the-art methods on split CIFAR-100, 5-datasets, and split ImageNet-R. Results of ESN are reported from the original paper~\cite{ESN}. Other results except for the upper-bounds are reported from DualPrompt~\cite{DUAL_PROMPT}.}
\label{table:main_result}
\end{table*} 

\section {Experiments}
Experimental section is start with the delineation of datasets and metrics employed. Then, we benchmark \ours in relation to state-of-the-art methodologies. Afterwards, we thoroughly studied each introduced component, affirming their effectiveness. The section is concluded by an analysis of \ours's efficiency from both memory and computational aspects.

\subsection {Datasets}
\noindent\textbf{Split CIFAR-100} is a standard continual learning benchmark wherein CIFAR-100 is partitioned into 10 discrete tasks. Extended results for alternative splits (e.g., 5 and 20 tasks) are available in the supplementary material.

\noindent\textbf{5-datasets} is a collection of datasets consisting of CIFAR-10~\cite{cifar10}, MNIST~\cite{mnist}, Fashion-MNIST~\cite{fashionmnist}, SVHN~\cite{svhn}, and notMNIST~\cite{notmnist}. Each is designated as a distinct task, mirroring real-world situations with significant differences between tasks.

\noindent\textbf{Split ImageNet-subset} is a frequently referenced benchmark that segments a 100-class subset of ImageNet~\cite{deng2009imagenet} into 10 tasks, each comprising 10 classes. It resembles challenging real-world image with high resolution.

\noindent\textbf{Split ImageNet-R} is first adapted for continual learning through DualPrompt~\cite{DUAL_PROMPT}. It seeks to replicate real-world conditions characterized by varied image styles and notable intra-class diversity. The original ImageNet-R~\cite{imagenetR} is divided into 24,000 training and 6,000 test images, with the 200 classes spread across 10 individual tasks.

\subsection {Configuration and Evaluation Metric}\label{sec:config}
\noindent\textbf{Configuration}. We use the following dataset-agnostic configuration unless stated otherwise. All experiments are conducted on four NVIDIA A100 GPUs. We train \ours for 50 epochs with a batch size of 256 using the AdamW optimizer~\cite{adamw}. The initial learning rate is set to $1 \times 10^{-3}$ and anneals to $1 \times 10^{-6}$ according to a cosine scheduler. The prompt length $L_p$ is set to 1 and deep prompt is used by default. The multi-centroid number $C$ and nearest neighbors $r$ are set to 5 and 3, respectively. A 3-layer MLP with 2048 hidden units and 768 output dimensions is randomly initialized at each new task. Other detailed configurations are provided in supplementary. 

\noindent\textbf{Evaluation metric}. We adopt the widely used average accuracy and forgetting from the end session~\cite{EMC, A_GEM, DUAL_PROMPT} as our evaluation metrics. We report average and standard deviation according to five runs with different random seeds. Detailed illustration of each metric and more results under alternative protocols are relegated to the supplementary.

\subsection {Benchmarking Against Leading Approaches} 
In this section, we first compare \ours to state-of-the-art methods that are compatible with the Transformer architecture on split-CIFAR100, 5-datasets, and split ImageNet-R datasets following DualPrompt~\cite{DUAL_PROMPT}. Then we reproduce state-of-the-art prototype-based methods as well as Transformer-based methods and compare \ours to them on split ImageNet-subset and split-CIFAR-100. Note that \textit{all methods reported in this section are implemented using the same pre-trained ViT-B/16.}

\begin{table*}[t]
\begin{center}
\resizebox{0.75\linewidth}{!}{%
\begin{tabular}{l c c cc cc cc}
	\toprule 
	\multirow{2}{*}{\textbf{Method}} &
	\multirow{2}{*}{\textbf{Buffer size}}&
	\multicolumn{3}{c}{\textbf{Split ImageNet-subset}} &
	\multicolumn{3}{c}{\textbf{Split CIFAR-100}} \\
	& & Backbone & Pretrain & Avg. Acc ($\uparrow$) & Backbone &  Pretrain & Avg. Acc ($\uparrow$) \\
	\midrule
	Upper-bound & - & ViT/B-16 & ImageNet & 94.22\scriptsize{$\pm$0.18} &
	ViT/B-16 & MAE & 93.15\scriptsize{$\pm$0.09} \\
	\midrule
	iCaRL~\cite{iCaRL} & 2000 & ResNet-18 & \xmark  &23.77\scriptsize{$\pm$0.35} & ResNet-18 & \xmark  & 51.12\scriptsize{$\pm$0.36}  \\
		
	PASS~\cite{PROTO_AUG} & 0 & ResNet-18 & \xmark & 27.16\scriptsize{$\pm$0.24} & ResNet-18 & \xmark & 36.32\scriptsize{$\pm$0.33} \\
				
	iCaRL~\cite{iCaRL} & 2000 & ViT-B/16 & MAE & 87.96\scriptsize{$\pm$0.26}  & ViT-B/16 & ImageNet & 75.10\scriptsize{$\pm$0.26}  \\
				
	PASS~\cite{PROTO_AUG} & 0 & ViT-B/16 & MAE & 72.72\scriptsize{$\pm$0.31} & ViT-B/16 & ImageNet & 64.10\scriptsize{$\pm$0.20} \\
				
	DualPrompt~\cite{DUAL_PROMPT} & 0 & ViT-B/16 & MAE & 92.50\scriptsize{$\pm$0.24} & ViT-B/16 & ImageNet & 86.51\scriptsize{$\pm$0.33} \\
				
	\bf \ours (ours) & 0 & ViT-B/16 & MAE &\bf 93.82\scriptsize{$\pm$0.06} & ViT-B/16 & ImageNet & \bf 91.12\scriptsize{$\pm$0.12} \\
	\bottomrule
    \end{tabular}
	}
\end{center}
\caption{Comparison to prototype-based methods and DualPrompt on split ImageNet-subset and split CIFAR-100. Results of DualPrompt on split CIFAR-100 are reported from the original paper. All other results are reproduced using the same pre-trained ViT-B/16.}
\label{table:proto}
\end{table*}

\noindent\textbf{Main results on split CIFAR-100, 5-datasets, and split ImageNet-R}. We compare \ours to regularization-based methods (\emph{EWC}~\cite{EMC} and \emph{LwF}~\cite{LwF}), advanced rehearsal-based methods ( \emph{ER}~\cite{ER}, \emph{GDumb}~\cite{GDumb}, \emph{BiC}~\cite{BiC}, \emph{DER++}~\cite{dark_memory}, and \emph{Co$^2$L}~\cite{Co2L}), and Transformer-based methods (\emph{L2P}~\cite{l2p}, \emph{ESN}~\cite{ESN}, and \emph{DualPrompt}~\cite{DUAL_PROMPT}). As shown in Table~\ref{table:main_result}, despite the rehearsal-free property of regularization-based methods, their results lack vigor. Rehearsal-based methods, on the other hand, produce decent results under a large memory budget. Yet, they are still outperformed by prompt-based methods, which do not require rehearsal. Among the family of new emerging Transformer-based methods, CPP surpasses alternatives by a significant margin, showcasing the superiority of our framework which leverages task-specific prompts that are optimized by the contrastive prototypical loss to steer prototypes.

\noindent\textbf{Comparison to prototype-based methods}.\label{sec:proto-related} We further compare \ours against state-of-the-art prototype-based methods including iCaRL~\cite{iCaRL} and PASS~\cite{PROTO_AUG} on split ImageNet-subset and split CIFAR-100. For calibration purpose, we also reproduce DualPrompt~\cite{DUAL_PROMPT} on split ImageNet-subset. \textit{To avoid information leakage, we adopt pretrained model using self-supervised MAE method for split ImageNet-subset}. Details on reproduction are consigned to the supplementary. As shown in Table~\ref{table:proto}, a pre-trained ViT backbone can significantly boost the performance of existing prototype-based methods, which aligns with the observation in \cite{ramasesh2022effect}. However, \ours still demonstrates an unparalleled performance with the same backbone, suggesting a clear advantage of \ours over alternatives.

 \begin{table*}[t]
	\begin{center}
		\resizebox{0.8\linewidth}{!}{%
			\begin{tabular}{l c c cc cc cc}
				\toprule 
				\multirow{2}{*}{\textbf{Pretrain}} & \multirow{2}{*}{\textbf{\ours}} & \multirow{2}{*}{\textbf{Multi-centroids}} & \multicolumn{2}{c}{\textbf{Split CIFAR-100}} & \multicolumn{2}{c}{\textbf{5-datasets}} & \multicolumn{2}{c}{\textbf{Split ImageNet-R}} \\
				& &  & Avg. Acc ($\uparrow$) & Forgetting ($\downarrow$) & Avg. Acc ($\uparrow$) & Forgetting ($\downarrow$) & Avg. Acc ($\uparrow$) & Forgetting ($\downarrow$) \\
				\midrule
				\multirow{4}{*}{Deit~\cite{deit}} & & & 71.9 & 9.97 & 70.54 & 0.28 & 51.29 & 5.84 \\
				& \checkmark & & 77.64\scriptsize{$\pm$0.18} & 8.06\scriptsize{$\pm$0.14} & 
				81.67\scriptsize{$\pm$0.13} & 1.71\scriptsize{$\pm$0.21} & 57.85\scriptsize{$\pm$0.07} & 6.74\scriptsize{$\pm$0.10} \\
				& & \checkmark & 74.23\scriptsize{$\pm$0.08} & 9.11\scriptsize{$\pm$0.14} & 72.36\scriptsize{$\pm$0.02} & \bf 0.11\scriptsize{$\pm$0.01} & 49.50\scriptsize{$\pm$0.16} & 6.16\scriptsize{$\pm$0.19} \\
				& \checkmark & \checkmark & \bf 82.24\scriptsize{$\pm$0.20} & \bf 6.05\scriptsize{$\pm$0.19} & \bf 90.94\scriptsize{$\pm$0.23} & 0.22\scriptsize{$\pm$0.04} & \bf 67.45\scriptsize{$\pm$0.25} & \bf 4.94\scriptsize{$\pm$0.12} \\
				\midrule
				\multirow{4}{*}{Dino~\cite{DINO}} & & & 76.69 & 8.91 & 72.18 & 0.68 & 45.59 & 7.77 \\
				& \checkmark & & 80.11\scriptsize{$\pm$0.22} & 6.88\scriptsize{$\pm$0.20} & 81.56\scriptsize{$\pm$0.02} & 0.74\scriptsize{$\pm$0.06} & 51.88\scriptsize{$\pm$0.10} & 8.67\scriptsize{$\pm$0.15} \\
				& & \checkmark & 79.71\scriptsize{$\pm$0.08} & 7.68\scriptsize{$\pm$0.07} & 74.31\scriptsize{$\pm$0.04} & \bf 0.18\scriptsize{$\pm$0.01} & 48.63\scriptsize{$\pm$0.18} & 5.72\scriptsize{$\pm$0.15} \\
				& \checkmark & \checkmark & \bf 83.59\scriptsize{$\pm$0.12} & \bf 5.43\scriptsize{$\pm$0.18} & \bf 89.10\scriptsize{$\pm$0.10} & 0.21\scriptsize{$\pm$0.08} & \bf 61.22\scriptsize{$\pm$0.59} & \bf 5.04\scriptsize{$\pm$0.08} \\
				\midrule
				\multirow{4}{*}{MAE~\cite{MAE}} & & & 74.65 & 8.60 & 72.77 & 0.30 & 55.25 & 6.21 \\
				& \checkmark & & 78.90\scriptsize{$\pm$0.32} & 8.42\scriptsize{$\pm$0.24} & 82.48\scriptsize{$\pm$0.06} & 
				0.41\scriptsize{$\pm$0.05} & 62.78\scriptsize{$\pm$0.09} & 5.88\scriptsize{$\pm$0.19} \\
				& & \checkmark & 76.68\scriptsize{$\pm$0.15} & 8.16\scriptsize{$\pm$0.09} & 73.98\scriptsize{$\pm$0.06} & 0.11\scriptsize{$\pm$0.04} & 54.68\scriptsize{$\pm$0.09} & 5.22\scriptsize{$\pm$0.20} \\
				& \checkmark & \checkmark & \bf 83.36\scriptsize{$\pm$0.18} & \bf 6.50\scriptsize{$\pm$0.38} & \bf 91.76\scriptsize{$\pm$0.02} & \bf 0.11\scriptsize{$\pm$0.01} & \bf 71.04\scriptsize{$\pm$0.30} & \bf 4.23\scriptsize{$\pm$0.07} \\
				\midrule
				\multirow{4}{*}{ViT~\cite{vit}} & & & 82.82 & 5.94 & 69.71 & 0.30 & 60.20 & 5.26 \\
				& \checkmark & & 87.18\scriptsize{$\pm$0.22} & 4.70\scriptsize{$\pm$0.33} & 81.54\scriptsize{$\pm$0.05} & 0.42\scriptsize{$\pm$0.05} & 65.88\scriptsize{$\pm$0.14} & 6.76\scriptsize{$\pm$0.02} \\
				& & \checkmark & 84.09\scriptsize{$\pm$0.05}  & 5.47\scriptsize{$\pm$0.12} & 72.1\scriptsize{$\pm$0.17} & \bf 0.15\scriptsize{$\pm$0.01} & 61.11\scriptsize{$\pm$0.23} & 4.84\scriptsize{$\pm$0.04} \\
				& \checkmark & \checkmark & \bf 91.12\scriptsize{$\pm$0.12} & \bf 3.33\scriptsize{$\pm$0.18} & \bf 92.92\scriptsize{$\pm$0.17} & 0.19\scriptsize{$\pm$0.07} & \bf 74.88\scriptsize{$\pm$0.07} & \bf 3.65\scriptsize{$\pm$0.03} \\
				\bottomrule
			\end{tabular}
		}
	\end{center}
	\caption{Ablation studies of the proposed modules under four different pre-training methods. When neither \ours nor multi-centroid prototype are applied, the model is reduced to the training-free baseline model as described in Sec.~\ref{sec:baseline}.}
	\label{table:main_ablation}
\end{table*} 

\subsection{Ablation Study}
\noindent\textbf{Effectiveness of the proposed modules}. Recognizing that the choice of embedding function is instrumental to our approach, it is critical to analyze \ours under different embedding functions. In pursuit of this, \ours is instantiated employing state-of-the-art pre-training mechanisms encompassing \emph{ViT}~\cite{vit}, \emph{Deit}~\cite{deit}, \emph{Dino}~\cite{DINO}, and \emph{MAE}~\cite{MAE}—a spectrum that spans supervised to self/un-supervised methodologies and discriminative to generative paradigms. As Table~\ref{table:main_ablation} elucidates, both task-specific prompts and multi-centroid prototypes remain robust across all pre-training methods, bringing around 10\% to 20\% absolute improvements over the baseline model. In addition, these modules, when isolated, retain efficacy and exhibit synergistic performance when combined together. An intriguing observation pertains to the variations induced by distinct pre-training paradigms, highlighting a pronounced correlation between baseline models and their \ours counterparts (\eg, $\rho=1.0$ for split CIFAR-100).

\begin{figure*}[t]
\begin{center}
	\includegraphics[height=0.125\textheight]{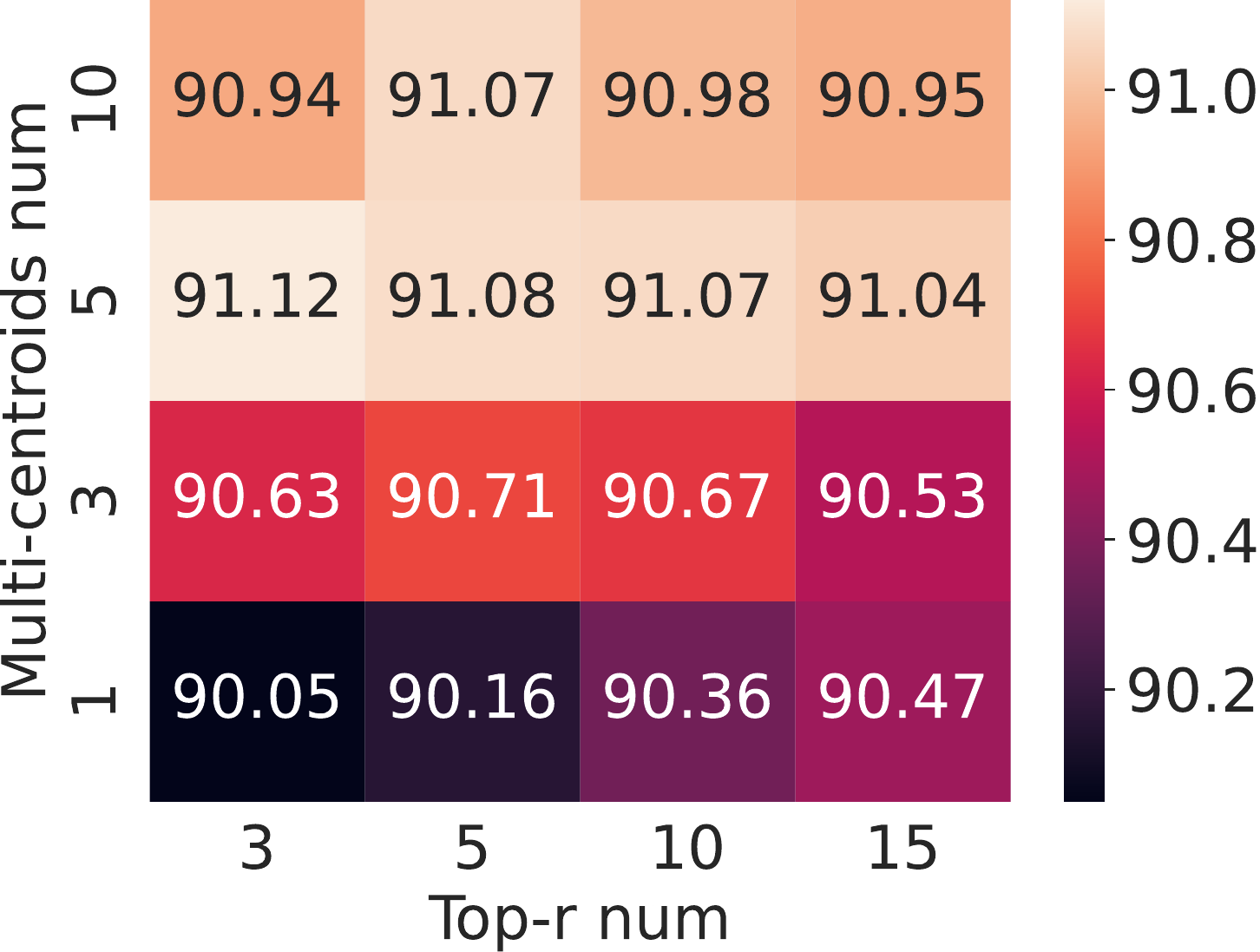}
	\hfill
	\includegraphics[height=0.125\textheight]{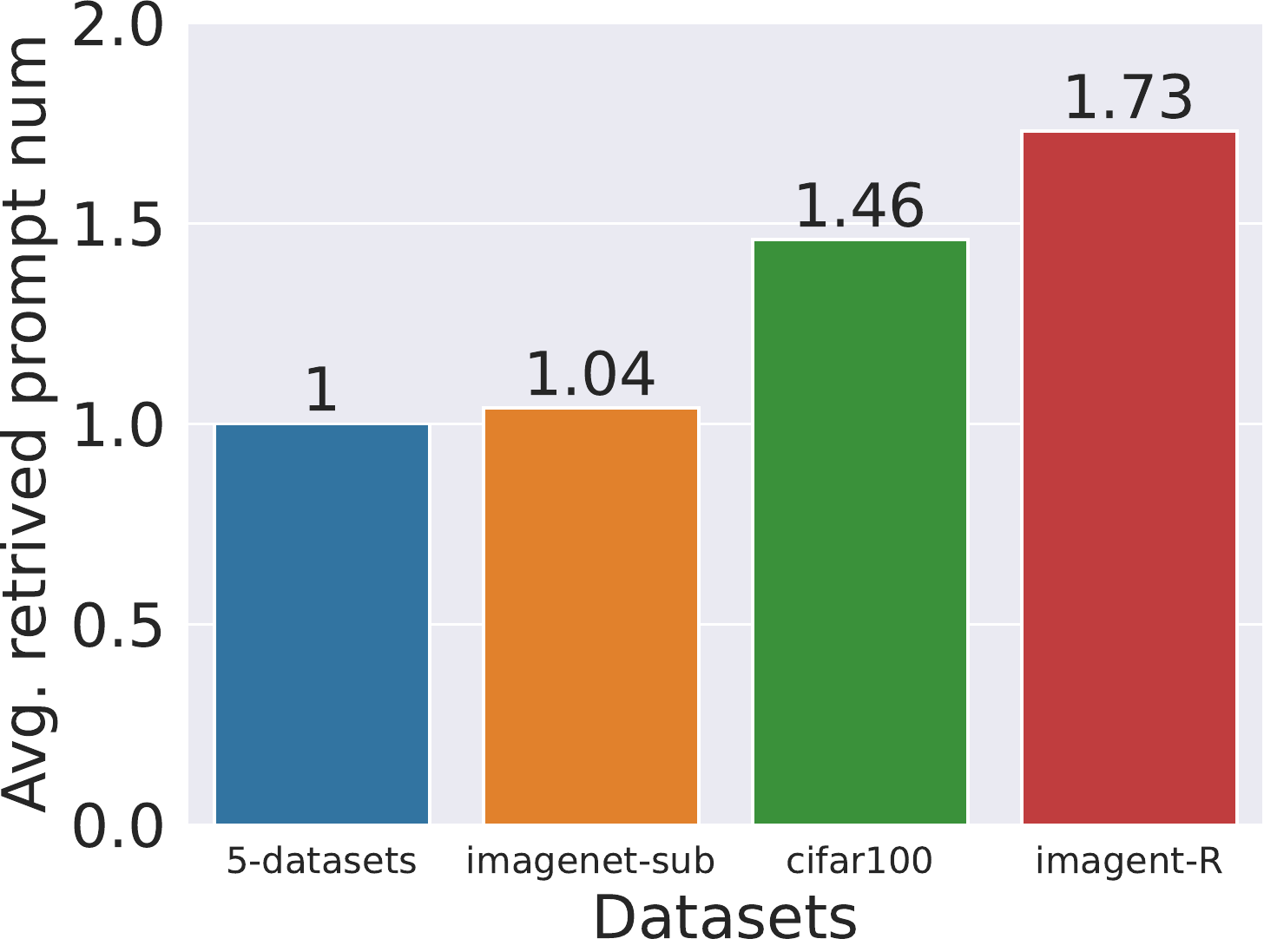}
	\hfill
	\includegraphics[height=0.125\textheight]{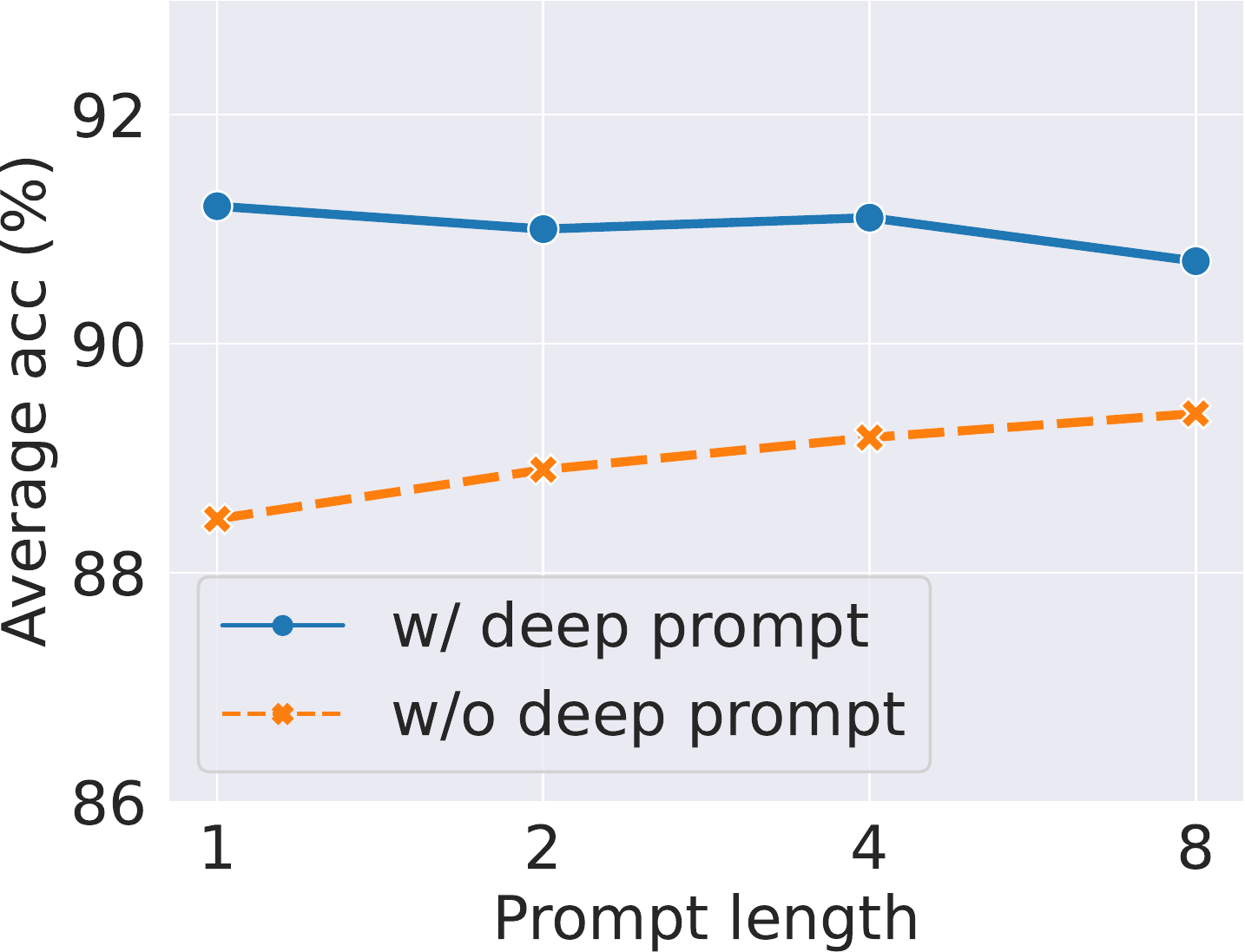}
    \hfill
	\includegraphics[height=0.125\textheight]{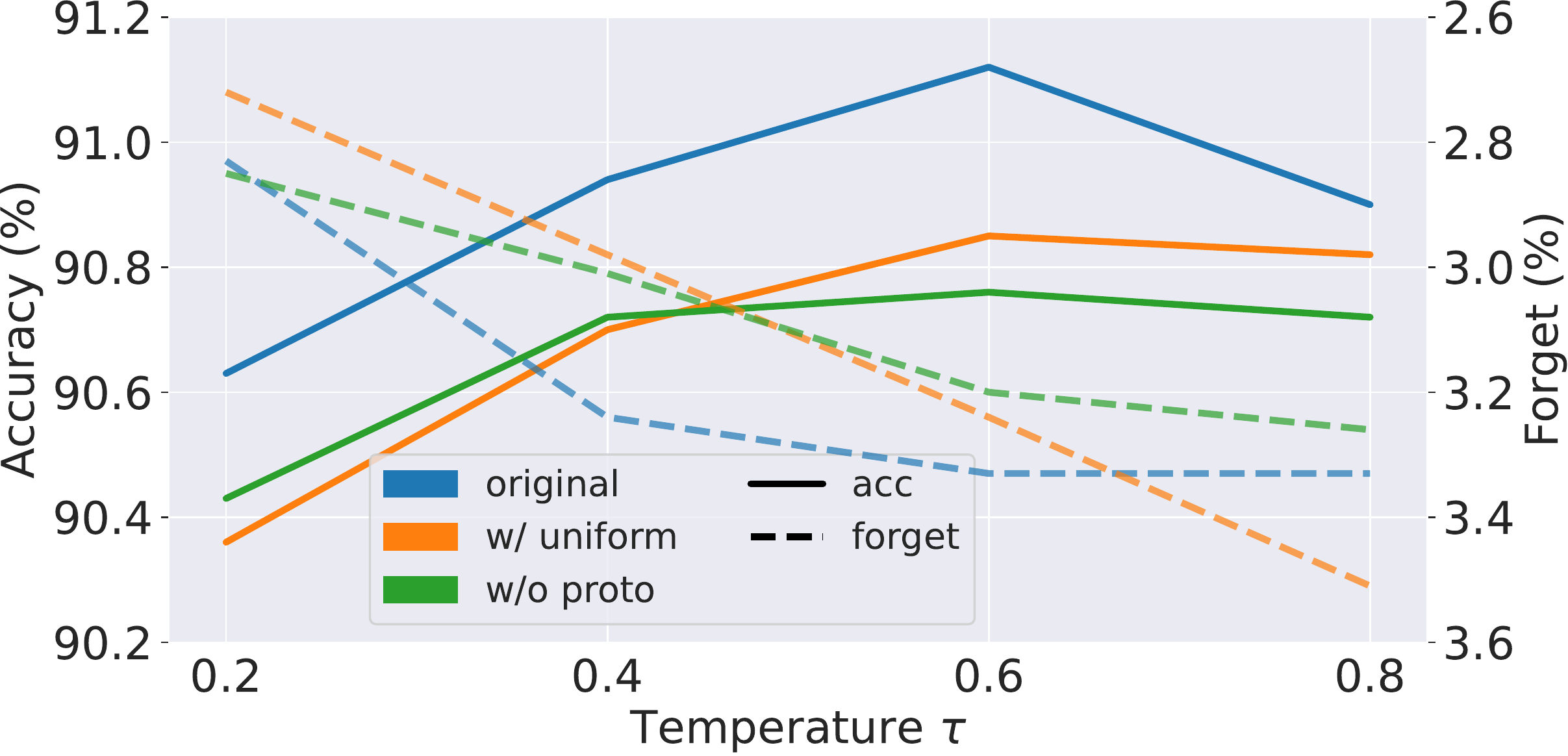}
\end{center}
	\caption{\textbf{Left:} Centroid number vs.\ query neighbors. \textbf{Middle left:} Our average number of retrieved prompts on different datasets. \textbf{Middle right:} Ablation studies on prompt length and deep prompt. \textbf{Right:} Results of CPL and its variants under different temperature coefficients.}
	\label{fig:4in1}
\end{figure*}

\noindent\textbf{Contrastive prototypical loss outperforms alternatives}. To discern the merits of \ourloss, we first compare it with two widely-used loss functions: \emph{CE} (cross-entropy) and \emph{SupCon} (supervised contrastive loss)~\cite{SCL}. As shown in Table~\ref{table:loss_ablation}, \ourloss outperforms both of them by a clear margin. Among comparisons, \emph{SupCon} is most compatible with ours, further confirming the advantage of coupling the contrastive loss design with a NCM classifier. Then, we independently add uniformity (w/ uniform) or remove prototypes (w/o proto), under varied temperatures to validate the efficacy of each proposed component. As shown in Fig.~\ref{fig:4in1} (right), encouraging uniformity decreases the performance, and removing prototypes aggravates the prototype interference, which are coherent with our analysis in Sec.~\ref{sec:cpl}.

\noindent\textbf{Stability and plasticity dilemma is mediated by temperature coefficient}.
Fig.~\ref{fig:4in1} (right) reveals the interplay between temperatures and model performance. Restrictive temperatures, which add greater penalties on negative pairs, reduce forgetting at the cost of damaging accuracy.  Higher temperatures, in opposite, improve the overall performance but escalating forgetting.  We empirically found $\tau=0.6$ makes a good trade-off between stability and plasticity and use it by default. 

\noindent\textbf{MLP is non-negligible}. We show in Table~\ref{table:loss_ablation} that non-linearity introduced by the MLP is crucial to the success training of prompts, irrespective of the underlying loss mechanism. Replacing the MLP neck by a single linear layer consistently leads to suboptimal outcomes.

\begin{table}[t]
	\begin{center}
		\resizebox{0.8\linewidth}{!}{%
			\begin{tabular}{l c c cc cc cc}
				\toprule 
				\multirow{2}{*}{\textbf{Method}} & \multicolumn{2}{c}{\textbf{Split CIFAR-100}} \\
				&  Avg. Acc ($\uparrow$) & Forgetting ($\downarrow$) \\
				\midrule
				CE (w/ linear neck) & 85.48\scriptsize{$\pm$0.49} & 8.42\scriptsize{$\pm$0.64}  \\
				SupCon (w/ linear neck)  & 87.94\scriptsize{$\pm$0.10} & 3.99\scriptsize{$\pm$0.10}  \\
				\ourloss (w/ linear neck) & 87.20\scriptsize{$\pm$0.12}& 4.08\scriptsize{$\pm$0.08} \\
				\midrule
				CE~\cite{CE} & 90.42\scriptsize{$\pm$0.04} &4.45\scriptsize{$\pm$0.17} \\
				SupCon~\cite{SCL} & 90.63\scriptsize{$\pm$0.27} & 3.35\scriptsize{$\pm$0.29}  \\
				\bf \ourloss (ours) &\bf 91.12\scriptsize{$\pm$0.12} & \bf 3.33\scriptsize{$\pm$0.18} \\
				\bottomrule
			\end{tabular}
		}
	\end{center}
	\caption{\ourloss vs.\ alternative loss functions.}
	\label{table:loss_ablation}
\end{table}

\noindent\textbf{Dissection of prompt architecture}. The architecture, comprising prompt length and depth, plays a pivotal role in dictating the efficacy of the task-specific prompt. As shown in Fig.~\ref{fig:4in1}~(middle right), deep prompts consistently outperforms their shallow counterparts, manifesting the importance of steering features at different levels of abstraction. Intriguingly, a configuration with $L_p=1$ when paired with deep prompts emerges as adequate for achieving commendable outcomes on split CIFAR-100, and excessive longer prompts can cause over-fitting.

\noindent\textbf{Centroid number vs.\ query radius}. The interplay between centroid number $C$ and query radius $r$ can impact the overall performance. To identify an appropriate configuration, we conduct a simple grid search on split CIFAR-100 and found the setting $C=5$ with $r=3$ works fairly well across benchmarks. As Fig.~\ref{fig:4in1}~(left) illustrates, the proposed multi-centroid strategy can better characterize the distribution of a given class distribution and thus effectively reduce the query radius and improve the performance.

\begin{figure}[t]
\begin{center}
	\includegraphics[width=0.4\linewidth]{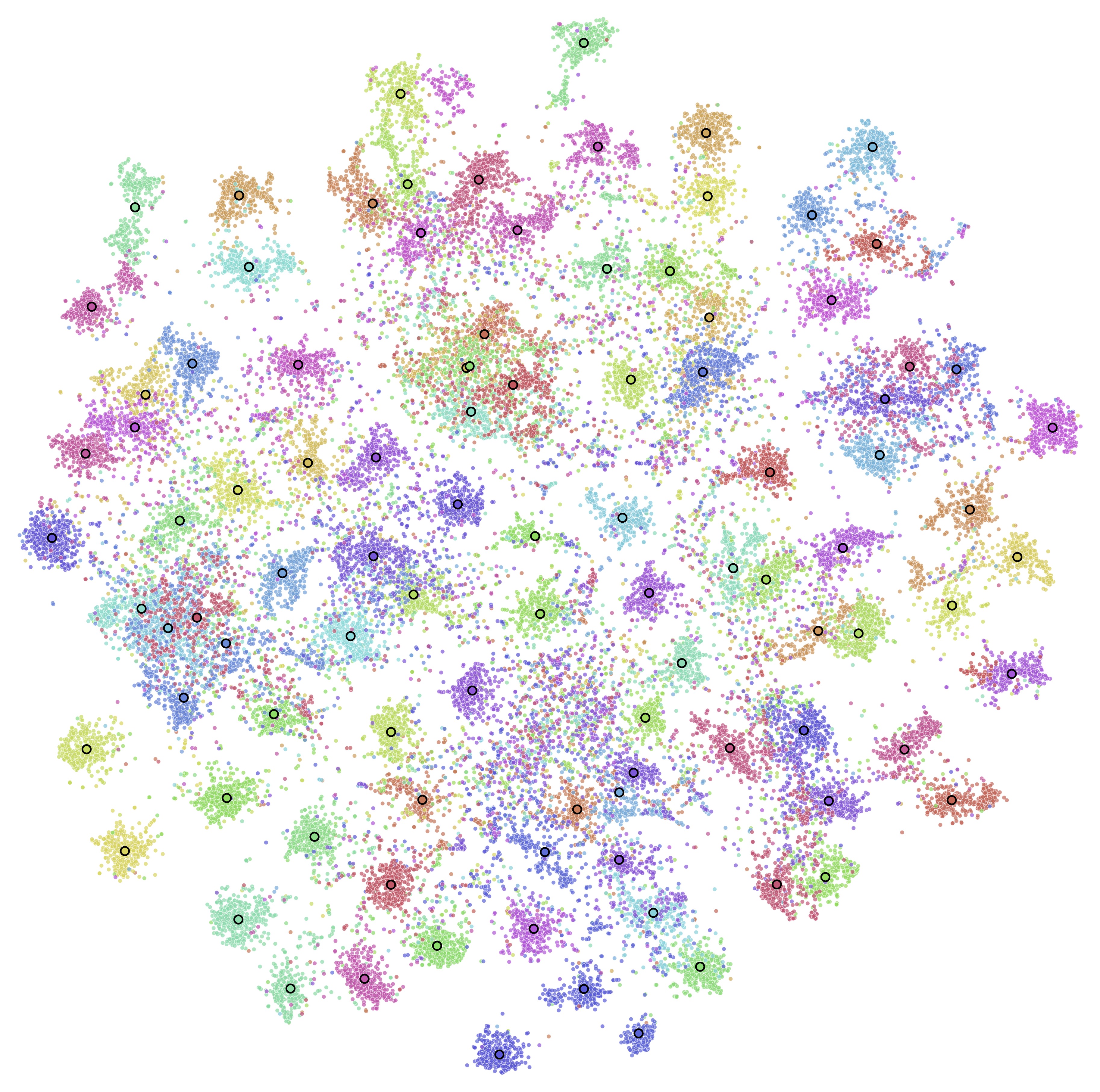}
	\hspace{0.4cm}
	\includegraphics[width=0.4\linewidth]{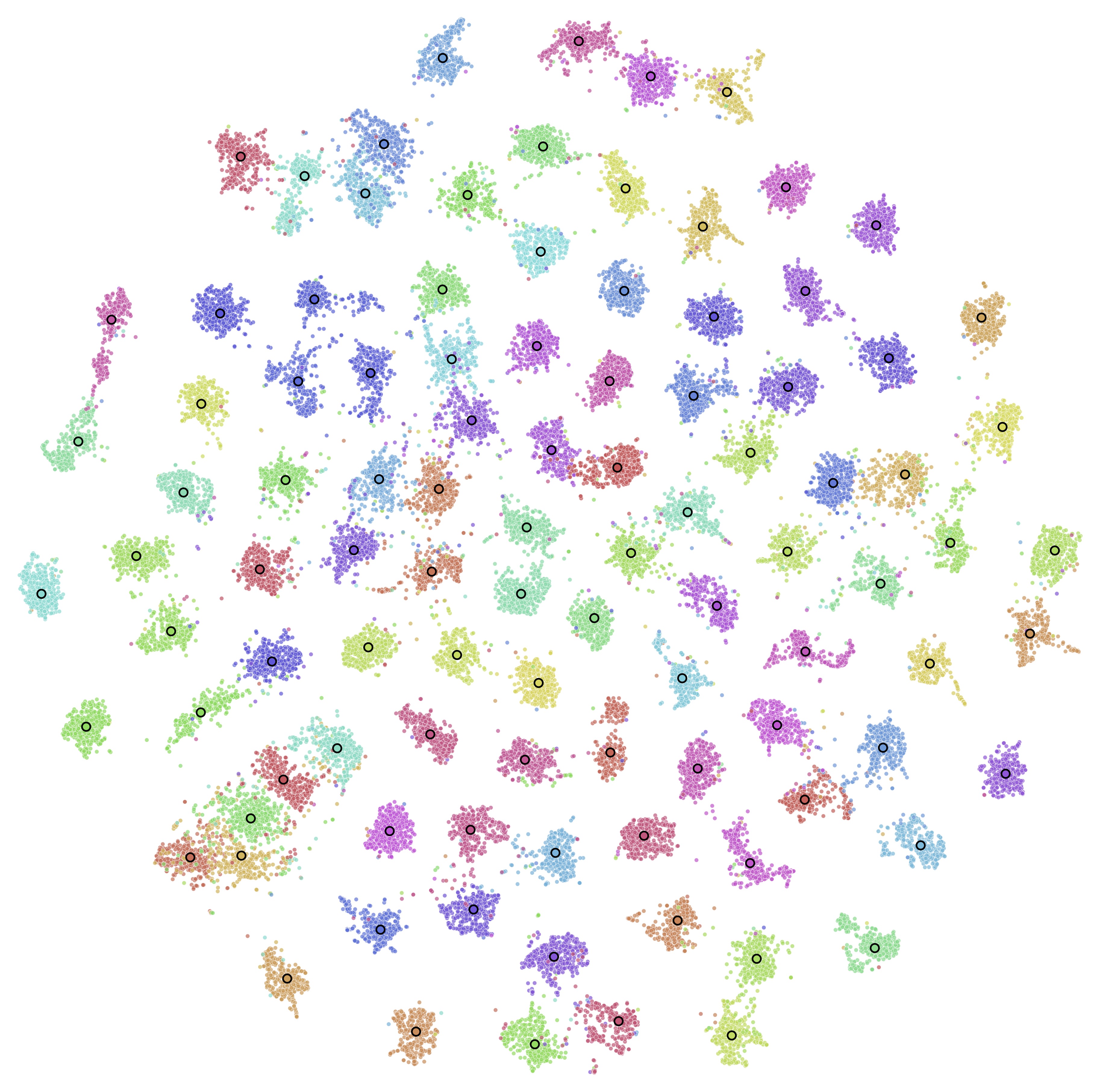}
\end{center}
\caption{t-SNE plots for data samples without (left) and with (right) task-specific prompts on CIFAR-100.}
\label{fig:visualization}
\end{figure}

\noindent\textbf{Visual effect of the task-specific prompts}. We visualize data instances alongside their corresponding prototypes from CIFAR-100, both in the presence and absence of task-specific prompts, in Fig.~\ref{fig:visualization}. Prior to the infusion of task-specific prompts, intra-class instances tend to cluster in the latent space,
Nonetheless, inter-class samples exhibit intertwined behavior. By introducing task-specific prompts, intra-class instances coalesce tightly, while disparate class clusters diverge markedly. We refer to supplementary for more visualization results and analysis.

\subsection{Analysis of Efficiency}
Here, we delve into the efficiency of \ours, examining memory consumption and computational demands. Comprehensive discussions are provided in the supplementary.

\begin{table}[t]
	\begin{center}
		\resizebox{0.8\linewidth}{!}{%
			\begin{tabular}{l c c cc cc cc}
				\toprule 
				\multirow{2}{*}{\textbf{Method}} & \multicolumn{2}{c}{\textbf{Split CIFAR-100}} \\
				&  Extra Mem. (MB / task) ($\downarrow$) &Avg. Acc  ($\uparrow$) \\
				\midrule
				DER++~\cite{dark_memory}  & 71.780 & 83.94\scriptsize{$\pm$0.34}\\
				L2P~\cite{l2p}  & 0.194 & 83.86\scriptsize{$\pm$0.28}\\
				DualPrompt~\cite{DUAL_PROMPT}  & 0.190 & 86.51\scriptsize{$\pm$0.33}\\
				\bf \ours (ours) & \bf 0.035 & \bf 91.12\scriptsize{$\pm$0.12} \\
				\bottomrule
			\end{tabular}
		}
	\end{center}
	\caption{Comparison on additional memory usage per task.}
	\label{table:memory_usage}
\end{table}

\noindent\textbf{Memory footprint}. For an in-depth understanding of memory efficiency, we contrast \ours against contemporaneous methods in the context of \textit{incremental memory consumption per task}. This metric underscores the growth in memory requirement with each additional task. Referencing Table~\ref{table:memory_usage}, \ours clearly distinguishes itself, registering a significantly minimized memory increment. This trait augments the scalability of \ours to handle long task sequences.

\noindent\textbf{Computational efficiency}. Thanks to the design of task-specific prompt and prototypes, the computational burden during training is modest: only tiny portion of parameters are updated through back-propagation and no explicit exemplars need to be forwarded. At inference, \ours may induce multiple forward propagations. This warrants an exploration of its computational feasibility.  As observed in Fig.~\ref{fig:4in1} (middle left), large inter-task divergences typically lead to fewer retrieved prompts (\eg, only one for 5-datasets), and thus have the similar inference cost as a linear classifier. As the intra-class divergence increases or inter-class divergence reduces, the retrieved prompt count amplifies. Yet, even on the challenging ImageNet-R dataset, CPP needs, on average, merely an additional $0.73$ forward passes, which is affordable under most scenarios.

\section{Conclusion}
We introduced an innovative, rehearsal-free framework for continual learning, capitalizing on task-specific prompts optimized through a tailored contrastive prototypical loss. This approach adeptly circumvents semantic drift and curtails prototype interference. The multi-centroid prototype strategy augments the expressiveness of prototypes. In evaluations, \ours outperforms established methodologies by  a large margin, with an in-depth assessment underscoring the salience of each component. We posit that \ours provides valuable insights into constructing scalable continual learning systems, informed by the recent advances in architectural design and representation learning. Future avenues include enhancing inference efficiency and broadening \ours to encompass complex scenarios such as blurred task boundaries, few-shot, and open-vocabulary configurations. Another promising direction involves devising automatic embedding function selection strategy so as to accommodate diverse downstream contexts.

{\small
\bibliographystyle{ieee_fullname}
\bibliography{egbib}
}

\onecolumn
\appendix

\section{Gradient Analysis of Contrastive Prototypical Loss}
\label{appendix:derivation}

Here, we provide an analysis of gradients for contrastive prototypical loss. To ease the notion, we abbreviate similarity between vector $\vz_i$ and $\vz_j$ as $s_{i,j}$ and denote negative set $N(i)\cup \hat{U}$ as $\hat{N}(i)$. Therefore, the loss of a data sample $\vx_i$ is:
\begin{equation}
	\label{eq:loss_for_k}
	\mathcal{L}_i = \frac{-1}{|P(i)|}\sum_{\vz_p \in P(i)}\log\frac{\exp(s_{i,p}/\tau)}{\underset{\vz_n \in \hat{N}(i)}{\sum}\exp(s_{i,n}/\tau)}.
\end{equation}
The gradient with respect to the similarity $s_{i,j}$ between a positive pair $(\vz_i, \vz_j)$ for $z_j \in P(i)$ can be derived as:
\begin{equation}
	\begin{aligned}
		\frac{\partial \mathcal{L}_i^k}{\partial s_{i,j}}
		&=\frac{-1}{|P(i)|}\sum_{\vz_p \in P(i)}\frac{\partial }{\partial s_{i,j}}\left(s_{i,p}/\tau - \log{\underset{\vz_n \in \hat{N}(i)}{\sum}}\exp(s_{i,n}/\tau)\right), \\
		& = \frac{-1}{|P(i)|}\sum_{\vz_p \in P(i)}\left(\frac{1}{\tau} \cdot \mathbbm{1}[p=j]-\frac{\frac{\partial}{\partial s_{i,j}}\left(\underset{\vz_n \in \hat{N}(i)}{\sum}\exp(s_{i,n}/\tau)\right)}{{\underset{\vz_n \in \hat{N}(i)}{\sum}}\exp(s_{i,n}/\tau)}\right), \\
		& = \frac{-1}{|P(i)|}\sum_{\vz_p \in P(i)}\left(\frac{1}{\tau}\cdot\mathbbm{1}[p=j]-0\right), \\
		& = \frac{-1}{\tau|P(i)|},
	\end{aligned}
\end{equation}
where $\mathbbm{1}$ is an indicator function. Similarly, the gradient with respect to the similarity $s_{i,m}$ between a negative pair $(\vz_i, \vz_m)$ for $\vz_m \in N(i)$ is: 
\begin{equation}
	\begin{aligned}
		\frac{\partial \mathcal{L}_i^k}{\partial s_{i,m}} & =\frac{-1}{|P(i)|}\sum_{\vz_p \in P(i)}\frac{\partial }{\partial s_{i,m}}\left(s_{i,p}/\tau - \log{\underset{\vz_n \in \hat{N}(i)}{\sum}}\exp(s_{i,n}/\tau)\right), \\
		& = \frac{-1}{|P(i)|}\sum_{\vz_p \in P(i)}\left(0-\frac{\frac{\partial}{\partial s_{i,m}}\left(\underset{\vz_n \in \hat{N}(i)}{\sum}\exp(s_{i,n}/\tau)\right)}{{\underset{\vz_n \in \hat{N}(i)}{\sum}}\exp(s_{i,n}/\tau)}\right), \\
		& = \frac{1}{|P(i)|}\sum_{\vz_p \in P(i)}\left(\frac{\underset{\vz_n \in \hat{N}(i)}{\sum}\left(\exp(s_{i,n}/\tau)\cdot \frac{1}{\tau} \cdot \mathbbm{1}[n=m]\right)}{{\underset{\vz_n \in \hat{N}(i)}{\sum}}\exp(s_{i,n}/\tau)}\right), \\
		& = \frac{1}{\tau} \frac{\exp(s_{i,m}/\tau)}{{\underset{\vz_n \in \hat{N}(i)}{\sum}}\exp(s_{i,n}/\tau)} .
	\end{aligned}
\end{equation}
We can see that, the gradient of the proposed loss follows the same pattern as the standard supervised contrastive loss. Positive pairs are treated equally and scaled by the temperature and the cardinality of the positive set. The property of implicit hard-case mining, \ie, proportional to the exponential term $\exp(s_{i,m}/\tau)$, is inherited from a typical contrastive loss in the negative term.

\section{\ours as an Energy-based Model}
\label{appendix:energy}
The overall objective of an energy-based model~\cite{LeCun06atutorial} (EBM) is to obtain an energy function $E_{\theta}(\vx): \R^{D}\rightarrow \R$ parameterized by $\theta$ that maps the high dimensional input $\vx$ to a scalar value. Giving an energy function $E_{\theta}(\cdot)$, its probability density $p(\vx)$ can be expressed through Gibbs distribution:
\begin{equation}
	p_{\theta}(y|\vx) = \frac{\exp(-E_{\theta}(\vx, y)/\tau)}{\int_{y'}\exp(-E_{\theta}(\vx, y')/\tau)} = \frac{\exp(-E_{\theta}(\vx, y)/\tau)}{\exp(-E_{\theta}(\vx)/\tau)},
	\label{eq:energy}
\end{equation}  
where $E_{\theta}(x)$ is the \textit{Helmholtz free energy} and $\tau$ is the temperature factor. Then we have:
\begin{equation}
	E_{\theta}(x) = \tau \cdot -\log\int_{y'}\exp(-E_{\theta}(\vx, y')/\tau).
\end{equation}
When making predictions under our framework, the categorical distribution can be represented as:
\begin{equation}
	p(y|\vx) = \frac{\exp(s_{x, y}/\tau)}{\sum_{y'=1}^{K}\exp(s_{x, y'}/\tau)},
	\label{eq:predict}
\end{equation}
where $s_{x, y} = \langle f_{\{\theta, P\}}(\vx), \hat{\vmu}_y\rangle$. When connecting Eq.~\ref{eq:predict} with  Eq.~\ref{eq:energy} and let $E_{\theta}(\vx, y) = -s_{x,y}$, we see that the energy of $\vx$ can be expressed as:
\begin{equation}
	E_{\theta}(\vx) =  \tau \cdot -\log\sum_{y=1}^{K}\exp(s_{x,y}/\tau),
\end{equation}
which is dominated by the largest similarity $s_{x,y^*}$ given an appropriate temperature $\tau$. The above analysis drives to a conclusion that assigning a data sample to its nearest prototype will generate the lowest energy for the system (\ie, a more stable system). Therefore, the question becomes whether the proposed contrastive prototypical loss serves as a qualified energy loss function. 

To see this, we first simplify Eq.~\ref{eq:loss_for_k} to a formula where there is only one positive sample $\vz_{p}$:
\begin{equation}
	\begin{aligned}
		\label{eq:sim_loss_for_k}
		\mathcal{L}_i &= -\log\frac{\exp(s_{i,p}/\tau)}{\underset{\vz_n \in \hat{N}(i)}{\sum}\exp(s_{i,n}/\tau)}, \\
		& = - s_{i,p} + \log{\underset{\vz_n \in \hat{N}(i)}{\sum}\exp(s_{i,n}/\tau)}.
	\end{aligned}
\end{equation}
By substituting $\vz_{p}$ with the target value prototype $\hat{\vmu}$, we have:
\begin{equation}
	\mathcal{L}_i = \underbrace{-\langle\vz_i, \hat{\vmu}\rangle}_\text{push down energy for target prototype} + \underbrace{\log{\underset{\vz_n \in \hat{N}(i)}{\sum}\exp(s_{i,n}/\tau)}}_\text{pull up energies for other prototypes}.
\end{equation}
When the above loss is minimized, the first term will push down the energy for target value prototype $\hat{\vmu}$ and the second term will increase energies for other prototypes. Hence, the above simplified loss is an effective loss function for the energy model. Note that the \textit{ground-truth} prototype $\hat{\vmu}$ is unavailable during training stage and it is approximated by a group of dynamically evolved positive embeddings from train data. It is worth pointing out that, unlike the contrastive divergence approximation used in \cite{li2020energy}, we also include previous classifiers in our loss. Nevertheless, our formulation does not suffer from the over suppression issue of previous classes likewise, as we use static non-parametric prototypes as classifiers. 

Finally, we show that encouraging uniformity is against the principle of the energy model. By encouraging uniformity as a typical supervised or self-supervised contrastive loss~\cite{SimCLR, SCL}, we rewrite Eq.~\ref{eq:sim_loss_for_k} into:
\begin{equation}
	\begin{aligned}
	\label{eq:with_uniform}
		\mathcal{L}_i &= -\log\frac{\exp(s_{i,p})}{\underset{\vz_n \in \hat{N}(i)}{\sum}\exp(s_{i,n}/\tau) + \underset{\vz_n\in\hat{P}(i)}{\sum}\exp(s_{i,n}/\tau)}, \\
		& = - s_{i,p} + \log\left({\underset{\vz_n \in \hat{N}(i)}{\sum}\exp(s_{i,n}/\tau)}+ \underbrace{\underset{\vz_n\in\hat{P}(i)}{\sum}\exp(s_{i,n}/\tau)}_\text{harmfully pull up energy for target prototype}\right),
	\end{aligned}
\end{equation}
where $\hat{P}(i) = \{z_j: y_j=y_i, i\neq j\}$. As shown in Eq~\ref{eq:with_uniform}, we can see that the second term in $\log$ operation acts adversely with respect to $-s_{i,p}$ which minimizes the energy between a sample and its corresponding prototype. Hence, pairs that encourage uniformity should be, in principle, removed from the negative set. Through the lens of energy-based models, we acknowledge that the training process of \ours is equivalent to minimize the energies of data samples and the inference process can be interpreted as selecting a task-specific prompt that minimizes the energy of a data sample in a given system.

\section{Training and Inference Algorithms}
\label{appendix:algorithm}
The training and inference algorithms of \ours are summarized in Algorithm~\ref{alg:train} and Algorithm~\ref{alg:test}, respectively. Note that when multi-centroid prototypes are used, the prototype generation process is substituted by a spectrum clustering process as described in Sec. 3.4.

\begin{algorithm}[t]
	\SetAlgoLined
	\textbf{Input:} Frozen embedding function $f_{\theta}$, number of tasks $T$,  training epochs $E$, training set ${\{(\vx_i^t, y_i^t)\}_{i=1}^{n_t}}\}_{t=1}^{T}$, prompt length $L_p$ and centriod number $C$. \\
	\For{$t = 1,\cdots,T$}{
		\textbf{Initialize:} MLP $m_{\sigma^t}$, task-specific group $P^t$, $U=\varnothing$, $\hat{U}=\varnothing$\\
		\For{class $k \in \mathcal{Y}^t$}{Generate key prototype $\vmu_k$ with Eq.1\\
			$U \leftarrow U \cup \vmu_k$}
		\For{$e = 1, \cdots, E$}{
			Optimize $\sigma^t$, $P^t$ through Eq.6
		}
		\For{class $k \in \mathcal{Y}^t$}
		{Generate value prototype $\hat{\vmu}_k$ with Eq.1 after prepending $P^t$\\
			$\hat{U} \leftarrow \hat{U} \cup \hat{\vmu}_k$}}
	\textbf{Output:} $\{P^t \}_{t=1}^T$, $U$ and $\hat{U}$
	\caption{Model Training}
	\label{alg:train}
\end{algorithm}

\begin{algorithm}[t]
	\SetAlgoLined
	\textbf{Given:} $f_{\theta}$, $U$, $\hat{U}$, $\{P^t\}_{t=1}^T$, query function $q(,,r)$. \\
	\textbf{Input:} Test image $\vx$\\ 
	\textbf{Initialize:}  $\hat{Q}=\varnothing$ \\ 
	$\vq = f_{\theta}(\vx)[0, :]$ \tcp*{Use class token as query vector.} 
	$ J = q(\vq, U, r)$ \tcp*{Retrieve indexes of J candidate prompts.}
	\For{$j \in J$}{
		$\vq^j = f_{\theta, P^j}(\vx)$ \\
		$\hat{Q} \leftarrow \hat{Q} \cup \vq^j$  \\
	}
	Make prediction following Eq.7\\
	\textbf{Output:} label $y$ 
	\caption{Model Inference}
	\label{alg:test}
\end{algorithm}

\section{Relation with Existing Taxonomy}\label{appendix:relation}
The field of continual learning features several classification schemas~\cite{PARISI201954, EmbracingChange}. In this discussion, we anchor our interpretation to the taxonomy presented in \cite{EmbracingChange}. At its core, our proposed method, \ours, represents a hybrid approach. When examining the task-specific prompt design, \ours echoes principles of modular architecture approaches~\cite{PNN, yoon2018lifelong, Li2019LearnTG}. These strategies typically augment learning capabilities in the face of novel tasks, ensuring task-specific refinements. Yet, \ours carves out its distinction by sidestepping the pitfalls of high memory consumption and computational demands. This efficiency is credited to the prompt-tuning mechanism we've adopted. Moreover, our method ingeniously deploys prototypes, which, when married with contrastively optimized task-specific prompts, avoid the requirement of task identities during inference. Assessing from the vantage point of these prototypes, \ours aligns with memory-based strategies~\cite{dark_memory, Co2L} and more pointedly with episodic memory techniques~\cite{episodic_memory}. Yet, while conventional memory strategies often rely on storing and revisiting explicit exemplars, \ours innovates by distilling these into compressed prototypes, which then double as classifiers. Lastly, from a regularization perspective, task-specific prompt-tuning inherently provide strong regularization in the parameter space by freezing most parameters. Drawing from the model functionality perspectives of \cite{FR}, our contrastive prototypical loss manifests as a functional space regulator. The prime constraint is to ensure minimal functional convergence between disparate class samples.

\section{Real-world Applicability of \ours}\label{appendix:usability}
The pragmatic viability of a continual learning approach is of great importance. We delve into the practicality of \ours by examining facets such as memory consumption, data privacy, and computational efficiency.

\noindent\textbf{Memory footprint}. It would be over-optimistic to generalize a model with the fixed learning capacity to ``life-long'' learning scenarios without forgetting. Thus, a judicious expansion strategy for the model is indispensable. Our proposal leverages prototypes with task-specific prompts to realize memory-efficient continual learning. Incremental memory demands in our scheme are essentially linked to storing these prototypes and task-specific prompts. For each novel class, \ours necessitates the addition of $12.2 \times 768$ parameters equating to roughly $1/15$ of a singular ImageNet image (with size $224\times224\times3$). Given such frugality, \ours holds promise for scaling, accommodating thousands of classes given a modern computational device.

\noindent\textbf{Data privacy}. Under privacy sensitive scenarios (\eg, healthcare applications), the retention of explicit data exemplars could be fraught with risks or outright impractical. In contrast to general memory-based methods, \ours encapsulates data as fictitious prototypes, offering an enhanced privacy shield. Therefore, \ours is arguably preferable for privacy sensitive scenarios.

\noindent\textbf{Computational efficiency}. During the learning phase, the prompt-tuning methodology ensures that back-propagation predominantly impacts only a tiny portion of parameters. As such, \ours is more computational efficient than methods that need to update all parameters under the same architecture. Moreover,
the knowledge retention in \ours is architectured around prototypes, used as reference anchors, obviating the need to forward raw exemplars. While the inference phase could impart some computational cost due to our task-specific prompt architecture, our empirical evaluations (as seen in Sec. 4.5) demonstrate that even on the most challenging benchmark, \ie, split ImageNet-R, this overhead remains affordable. Crucially, the granularity of this computational overhead can be fine-tuned by modulating the centroid number, $C$, and the neighbor radius, $r$.

\section{Experimental Setups}
\subsection{Evaluation Metrics}\label{appendix:metrics}
Let $A_{i, j}$ denotes the classification accuracy for the $j$-th task after training on the $i$-th task. The Average Accuracy ($A_i$) and Forgetting ($F_i$) aftering learning task $i$ is then defined as follows:
\begin{equation*}
	\begin{aligned}
		&A_{i}=\frac{1}{i} \sum_{j=1}^{i} A_{i, j},\\
		&F_{i}=\frac{1}{i-1} \sum_{j=1}^{i-1} \max _{j^{\prime} \in\{1, \cdots, i-1\}}\left(A_{j^{\prime},j}-A_{i,j}\right).
	\end{aligned}
\end{equation*}
Assuming there are $T$ tasks in total, we report the accuracy from the end session as $Acc=A_T$ following \cite{EMC, DUAL_PROMPT}. We notice that some previous studies~\cite{LwF, PROTO_AUG, dytox} report the macro average over all sessions, \ie, $Acc=\frac{1}{T}\sum_{i=1}^T A_i$. To ease reference, we provide results under both protocols in Appendix~\ref{appendix:multi_protocal}.

\subsection{Data Augmentations}\label{appendix:aug}
In this section, we detail transformations used in our experiments. In general, we adopt standard data transformations used for contrastive learning, which include:
\begin{itemize}
	\item RandomResizedCrop (size$=224$, scale$=[0.8, 1.0]$)
	\item RandomHorizontalFlip ($p=0.5$)
	\item RandomColorJitter ($p=0.8$, brightness$=0.4$, contrast$=0.4$, saturation$=0.2$, hue$=0.1$)
	\item RandomGrayscale ($p=0.2$)
	\item RandomGaussianBlur ($p=0.1$, min radius$=0.1$, max radius=2)
	\item RandomSolarization ($p=0.2$)
	\item Normalization
\end{itemize}
Note that the normalization operation of an embedding function uses the same statistics as the ones used in its corresponding pre-training method.

\section{Implementation Details for Reproducibility}\label{appendix:reproduce}
\begin{itemize}
\item \noindent\textbf{Upper-bound}. For establishing upper-bound performance, we fine-tune the pre-trained embedding functions on the target datasets with all classes visible. The initial learning rate is set to $1 \times 10^{-4}$. All other settings follow the same configuration as training \ours to keep a fair comparison. One exception is fine-tuning MAE on ImageNet-subset where we found the transformations tailored for \ours generate inferior results. As such, we have adhered to the data transformations detailed in the original work~\cite{MAE}.

\item \noindent\textbf{iCaRL}. For iCaRL, we initiate our experimentation using the ResNet-18 model, following the original setting~\cite{iCaRL}. Given that the choice of the embedding function is orthogonal to the technical design in iCaRL, we switch the embedding function to the same pre-trained Transformer as used in \ours. The learning rate is set to $1 \times 10^{-4}$ according to a grid search and the memory size is set to 2000. 

\item \noindent\textbf{PASS}. For PASS~\cite{PROTO_AUG}, the original paper uses 50 classes in the initial task and 5 class for each incremental task. To synchronize with our configuration, we modify the split to 10 classes per task and 10 tasks in total. Similar to reproducing iCaRL, we run experiments under both ResNet-18 and pre-trained Transformer backbones and the learning rate is set to $1 \times 10^{-4}$ for Transformer. Other hyper-parameters remain the same as in the original paper. 
 
\item \noindent\textbf{DualPrompt}. DualPrompt~\cite{DUAL_PROMPT} is originally built upon Transformer structures, so we only change the pre-trained weights to MAE to avoid information leakage. All other parameters adhere strictly to the guidelines of the originating paper. Specifically, we set $L_e=20$, $L_g=5$, $start_e=3$, $end_e=5$, $start_g=1$, and $end_g=2$. We train the model for 50 epochs with the constant learning rate of $0.005$ and the Adam optimizer is used.
\end{itemize}

\section{Additional Empirical Results}
\subsection{Results under Different Protocols}\label{appendix:detailed_result}

\begin{figure*}[t]
\begin{center}
	\includegraphics[width=0.32\textwidth]{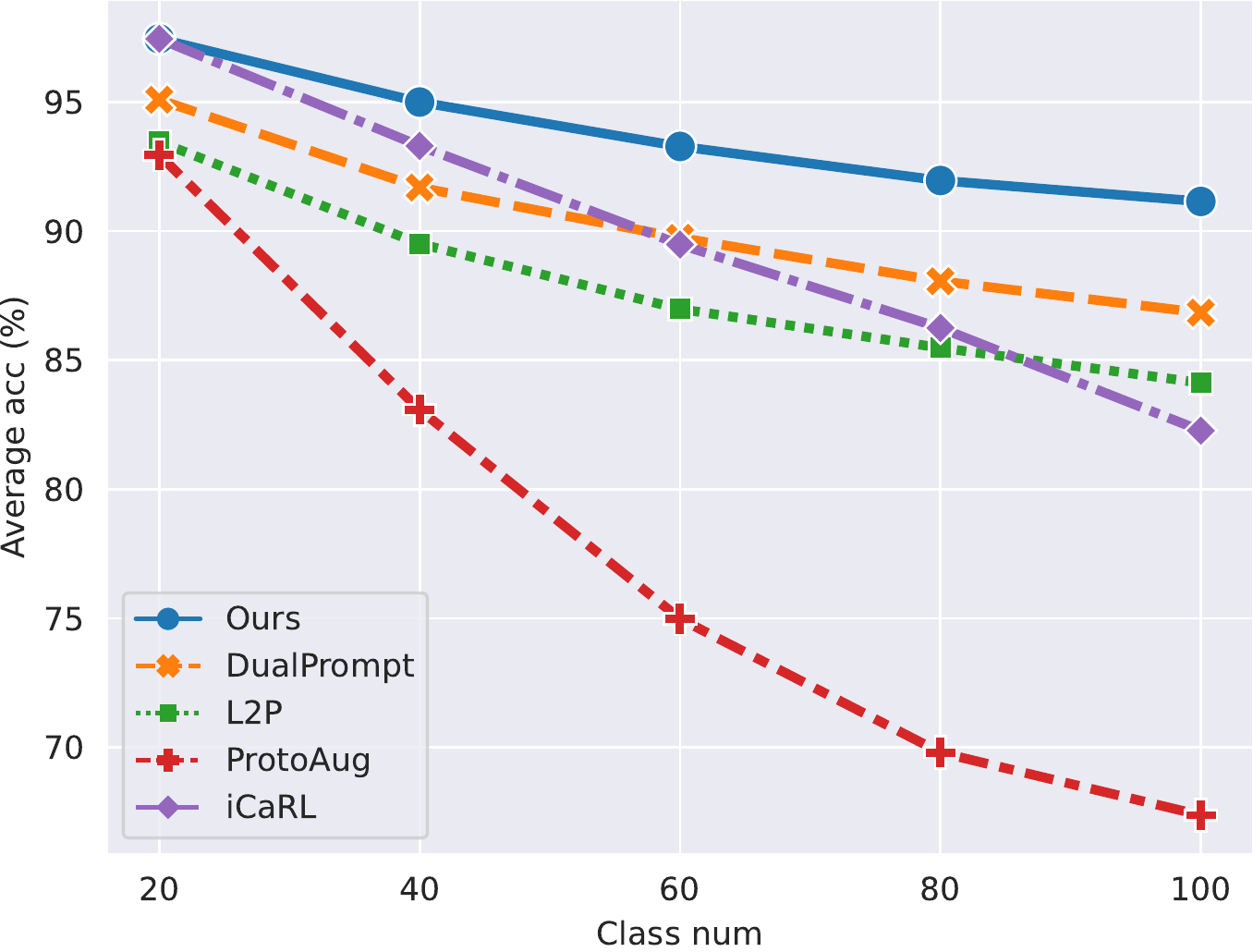}
	\hfill
	\includegraphics[width=0.32\textwidth]{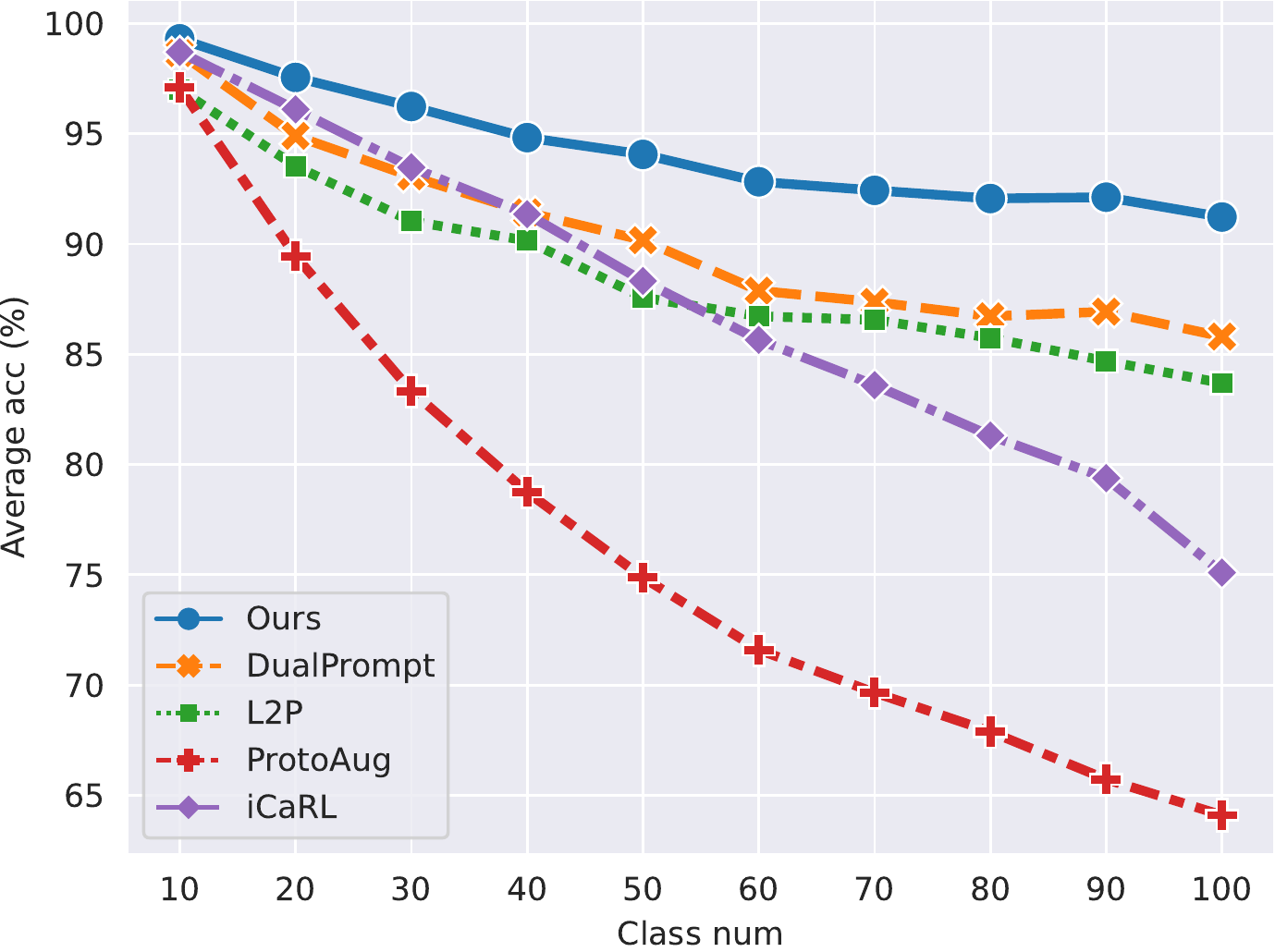}
	\hfill
	\includegraphics[width=0.32\textwidth]{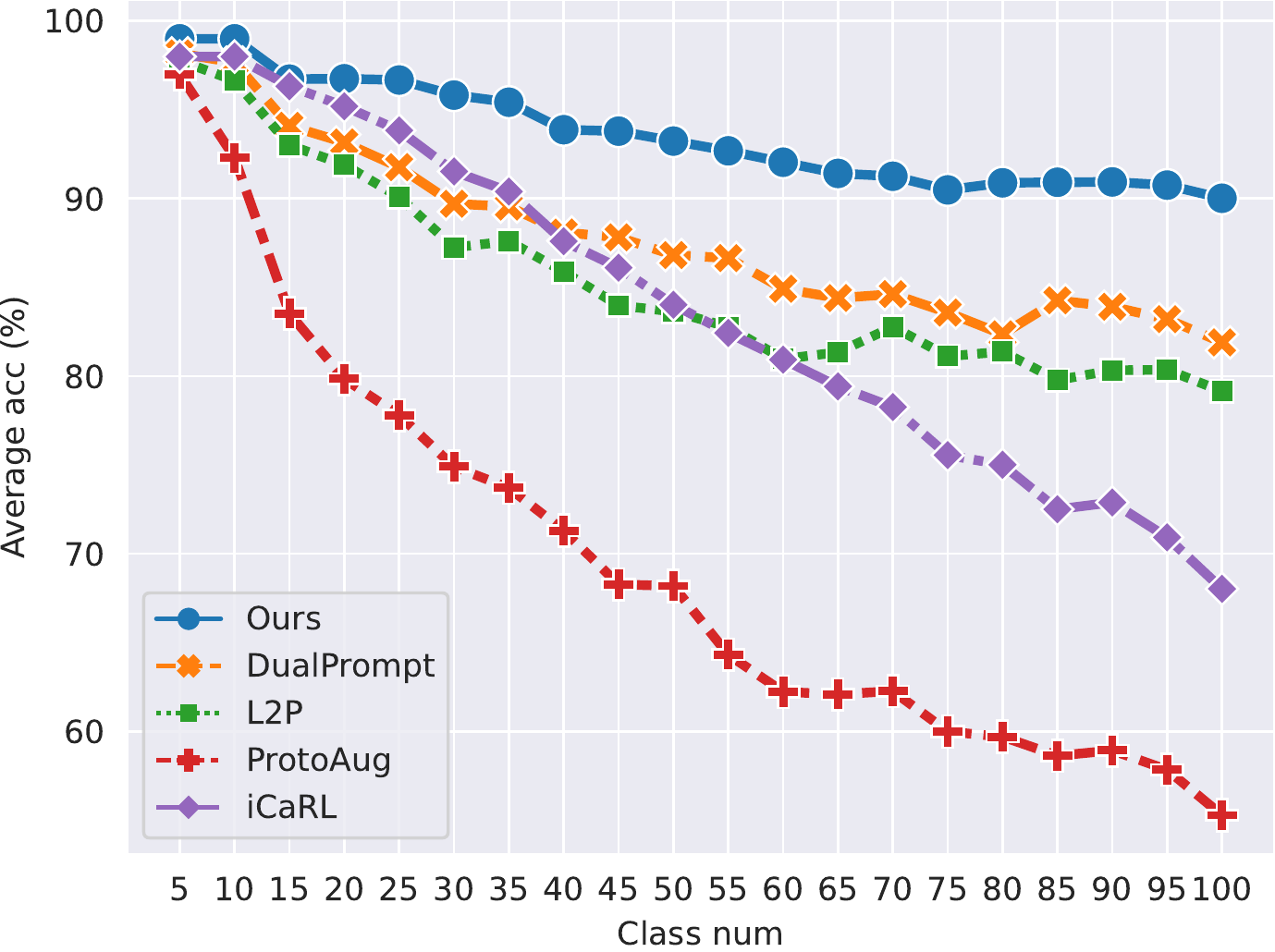}
\end{center}
\caption{Comparison with state-of-the-art methods on CIFAR-100 under multiple splits. \textbf{Left:} 5-splits, \textbf{Middle:} 10-splits \textbf{Right:} 20-splits.}
\label{fig:multi_splits}
\end{figure*}

\noindent\textbf{Detailed results on CIFAR-100 under different splits}. Here, we provide detailed task-wise results on split CIFAR-100 under different splits. As shown in Fig.~\ref{fig:multi_splits}, \ours exhibits clear and consistent improvements over other methods and the gaps are enlarged as the length of the task sequence increases.

\noindent\textbf{Results under different metric}.\label{appendix:multi_protocal} In Table~\ref{table:diff_protocol}, we provide results under two different protocols as described in Appendix~\ref{appendix:metrics}.

\subsection{Comparison to Architectural Methods}
In this section, we compare \ours with different architectural methods. It is non-trivial to migrate ConvNet-based methods to Transformer-based methods, so we follow the practice in DualPrompt~\cite{DUAL_PROMPT} to measure the difference between a method and its own upper-bound. As shown in Table~\ref{table:architecture}, \ours largely bridges the gap between incremental learning and joint learning. Besides, \ours is also more memory efficient than alternatives. 

\begin{table}[t]
	\begin{center}
	\resizebox{0.7\linewidth}{!}{%
			\begin{tabular}{c c c cc cc cc}
				\toprule 
				\multirow{2}{*}{\textbf{Task num}} &
				\multirow{2}{*}{\textbf{Dataset}} &
				\multirow{2}{*}{\textbf{Pre-train}} &
				\multicolumn{2}{c}{\textbf{Accuracy}} &
				\multicolumn{2}{c}{\textbf{Forgetting}}\\
				& &  & Avg. ($\uparrow$) & Last ($\uparrow$) & Avg. ($\downarrow$) & Last ($\downarrow$)\\
				\midrule
				
				5 & split CIFAR-100 & ViT & 
				93.78\scriptsize{$\pm$0.12} & 91.20\scriptsize{$\pm$0.06} & 2.77\scriptsize{$\pm$0.18} & 2.98\scriptsize{$\pm$0.21} \\
				
				10 & split CIFAR-100 & ViT & 94.18\scriptsize{$\pm$0.09} & 91.12\scriptsize{$\pm$0.12} & 2.39\scriptsize{$\pm$0.22} & 3.33\scriptsize{$\pm$0.18} \\
				
				20 & split CIFAR-100 & ViT & 93.49\scriptsize{$\pm$0.07} & 89.81\scriptsize{$\pm$0.17} & 2.84\scriptsize{$\pm$0.15} & 3.70\scriptsize{$\pm$0.13} \\
				
				5 & 5-datasets & ViT & 
				94.77\scriptsize{$\pm$0.19} & 
				92.92\scriptsize{$\pm$0.17} & 
				0.16\scriptsize{$\pm$0.08} & 
				0.19\scriptsize{$\pm$0.07} \\
				
				10 & split ImageNet-Sub & MAE & 95.14\scriptsize{$\pm$0.06} & 93.82\scriptsize{$\pm$0.06} & 0.82\scriptsize{$\pm$0.04} & 1.98\scriptsize{$\pm$0.06} \\
				
				10 & split ImageNet-R & ViT & 78.22\scriptsize{$\pm$0.14} & 74.88\scriptsize{$\pm$0.07} & 3.32\scriptsize{$\pm$0.19} & 3.65\scriptsize{$\pm$0.03} \\
				\bottomrule
		\end{tabular}
	}
	\end{center}
	\caption{Results of \ours under different protocols.}
	\label{table:diff_protocol}
\end{table}

\begin{table}[t!]
		\begin{center}
			\resizebox{0.85\linewidth}{!}{%
			\begin{tabular}{lclccc>{\centering\arraybackslash}p{1.8cm}>{\centering\arraybackslash}p{1.8cm}}
					\toprule 
					\multirow{2}{*}{\textbf{Method}} & \multirow{2}{*}{\textbf{Backbone}} & \multirow{2}{*}{\textbf{Avg. Acc ($\uparrow$)}} & \multirow{2}{*}{\textbf{Diff ($\downarrow$)}} & \multirow{2}{*}{\textbf{Pretrained}} &
					\multirow{2}{*}{\textbf{Buffer size}} &  \multicolumn{2}{c}{\textbf{Additional Parameters}} \\
					& & & & & & MB & \% \\
					\midrule
					Upper-bound & \multirow{5}{*}{ResNet18}& 80.41$^\dagger$ & - & - & - & - & - \\
					SupSup~\cite{supsup} & & 28.34\scriptsize{$\pm$2.45}$^\ddagger$ & 52.07 & \xmark & 0 & 3.00 & 6.50\% \\
					DualNet~\cite{DualNet} & & 40.14\scriptsize{$\pm$1.64}$^\ddagger$ & 40.27 & \xmark & 1000 & 5.04 & 10.90\% \\
					RPSNet~\cite{RPSNet} & & 68.60$^\dagger$ & 11.81 & \xmark & 2000 & 181.00 & 40.4\% \\
					DynaER~\cite{DynaER} & & 74.64$^\dagger$ & 5.77 & \xmark & 2000 & 19.80 & 43.80\% \\
					\midrule
					Upper-bound & \multirow{2}{*}{ResNet152}& 88.54$^\dagger$ & - & - & - & - & - \\
					DynaER~\cite{DynaER} & & 71.01\scriptsize{$\pm$0.58}$^\ddagger$ & 17.53 & \xmark & 2000 & 159.00 & 68.50\% \\
					\midrule
					Upper-bound & \multirow{2}{*}{Customized ViT}& 76.12$^\dagger$ & - & - & - & - & - \\
					DyTox~\cite{dytox} & & 62.06\scriptsize{$\pm$0.25}$^\dagger$ & 14.06 & \xmark & 2000 & 0.04 & 0.38\% \\
					\midrule
					Upper-bound & \multirow{4}{*}{ViT-B/16}& 93.15\scriptsize{$\pm$0.09} & - & -  & - & - & - \\
					L2P~\cite{l2p} & & 83.86\scriptsize{$\pm$0.28}$^\ddagger$ & 9.29 & \checkmark & 0 & 1.94 & 0.56\% \\
					DualPrompt~\cite{DUAL_PROMPT} & & 86.51\scriptsize{$\pm$0.33}$^\ddagger$ & 6.64 & \checkmark & 0 & 1.90 & 0.55\%  \\
					\bf \ours (ours) & & \bf 91.12\scriptsize{$\pm$0.12} & \bf 2.03 & \checkmark & 0 & \bf 0.35 & \bf 0.10\%  \\
					\bottomrule
				\end{tabular}
			}
		\end{center}
    \caption{Comparison with architecture-based methods on split CIFAR-100. \texttt{Diff} (lower is better) measures performance gap between a method and its corresponding upper-bound under the same backbone. $^\dagger$ denotes the results reported from the original papers. $^\ddagger$ represents results copied from DualPrompt~\cite{DUAL_PROMPT}.}
    \label{table:architecture}
\end{table}

\subsection{Extra Ablation Studies}\label{appendix:extra_ablation}
\begin{figure}[t!]
    \centering
    \begin{minipage}{0.4\textwidth}
        \centering
        \includegraphics[height=0.7\textwidth]{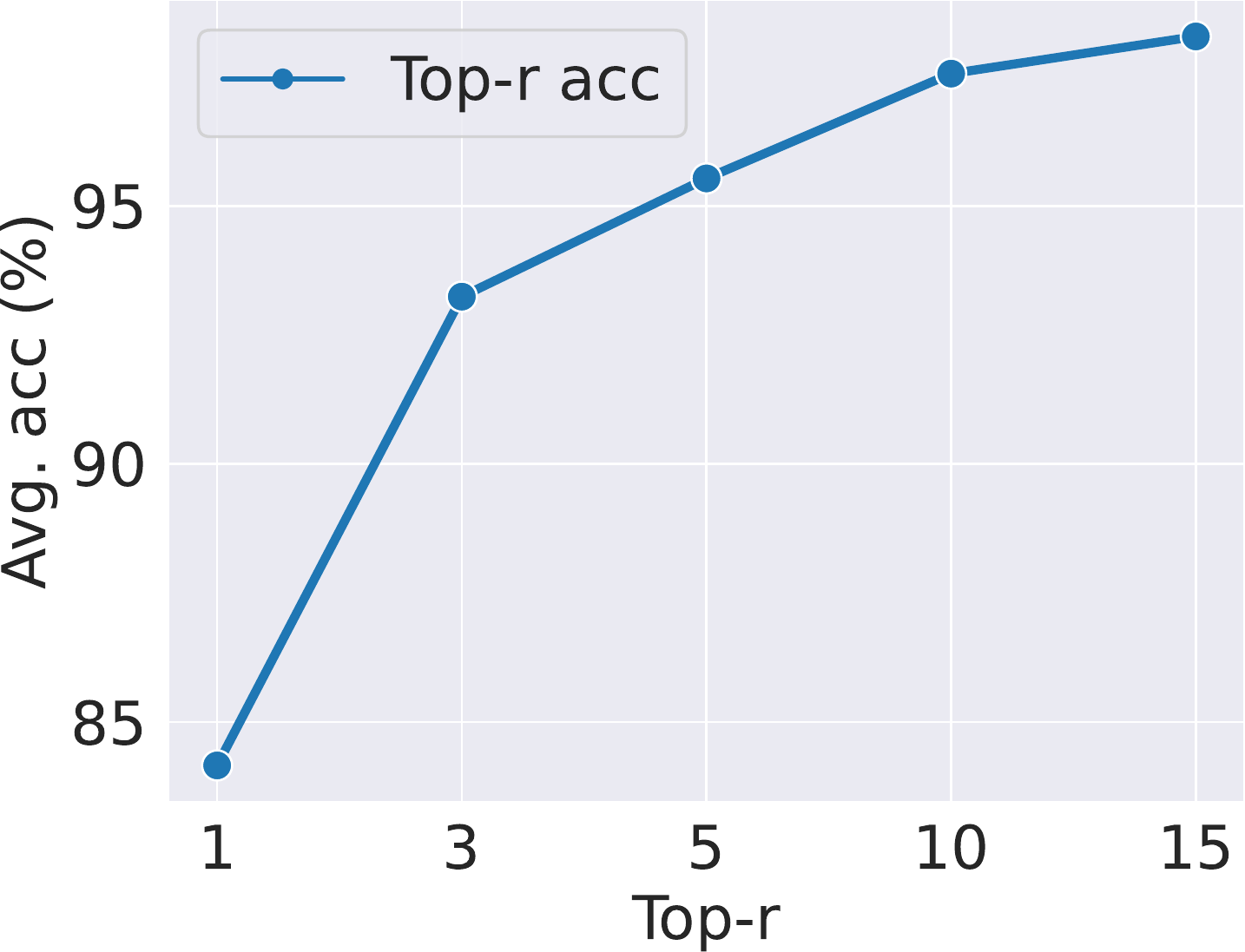} 
        \caption{Top-r retrieval accuracy on split CIFAR-100 using 5-centroid prototype.}
        \label{fig:top-r}
    \end{minipage}
    \hspace{1cm}
    \begin{minipage}{0.4\textwidth}
        \centering
        \includegraphics[height=0.7\textwidth]{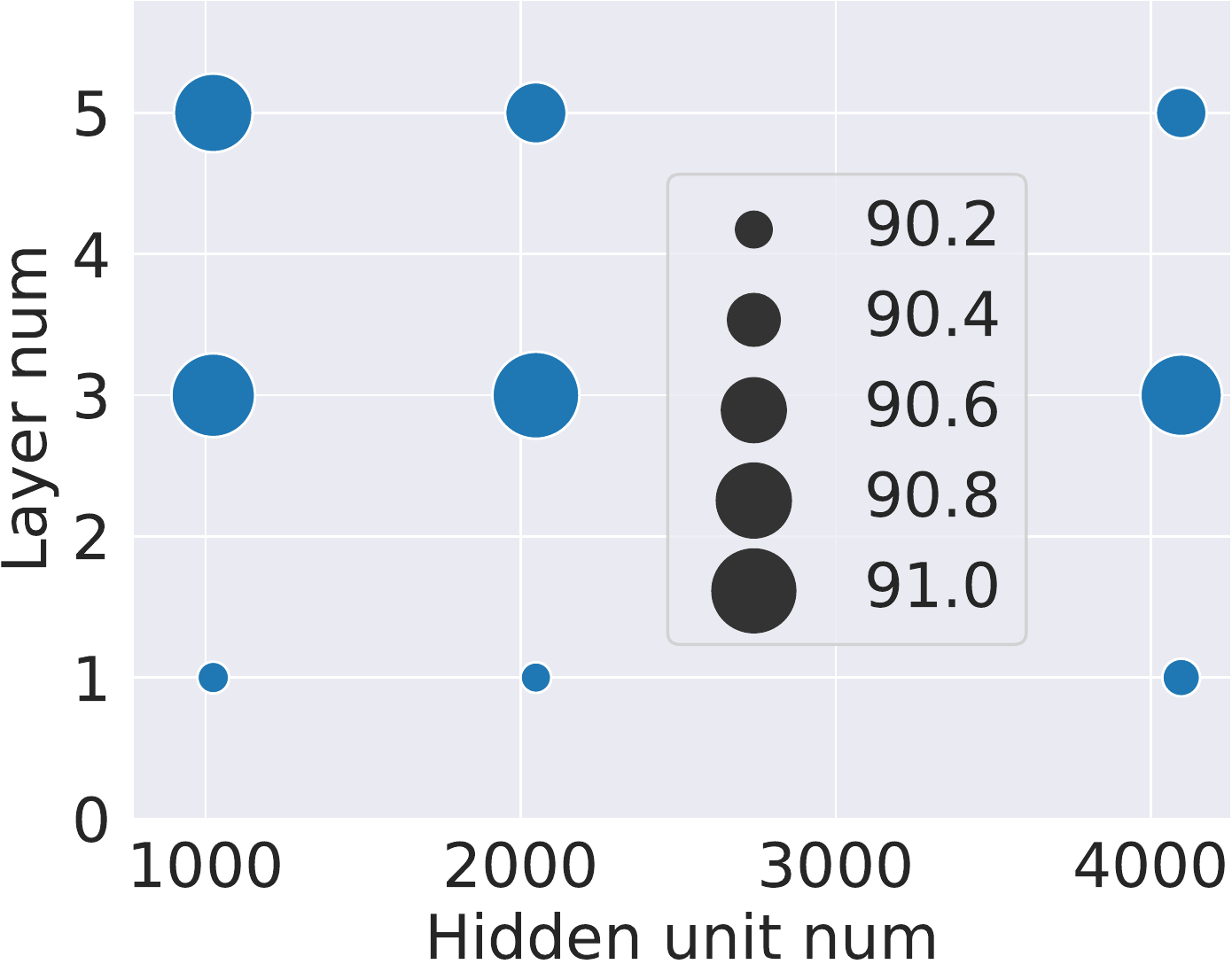} 
        \caption{Ablation studies on the layer number and hidden units of MLP neck.}
        \label{fig:mlp}
    \end{minipage}
\end{figure}

\noindent\textbf{Ablation study on MLP designs}. As the MLP neck is important for training task-specific prompts in our framework, we here probe relations between the MLP width (the number of hidden units), depth (the number of layers), and prompt quality. As shown in Fig.~\ref{fig:mlp}, either monotonously increasing the number of layers or hidden units does not bring benefits. A three-layer MLP with 2048 hidden units, which is the same as 
the conventional MLP neck used in self-supervised representation learning, produces the best performance. Therefore, we adopt this setting as default across our experiments.

\noindent\textbf{Generation of multi-centriod prototypes}. In \ours, we leverage spectral clustering to produce multi-centroid prototypes. Herein, we provide results using the commonly used k-means clustering algorithm. As shown in Table~\ref{table:clustering}, spectral clustering empirically demonstrates a better performance. 

\begin{table}[t]
	\begin{center}
		\resizebox{0.35\linewidth}{!}{%
			\begin{tabular}{c c c cc cc cc}
				\toprule 
				\multirow{2}{*}{\textbf{Methods}} &
				\multicolumn{2}{c}{\textbf{Split CIFAR-100}} \\
				& Avg. Acc ($\uparrow$) &  Forget ($\downarrow$) \\
				\midrule
				K-means & 90.77\scriptsize{$\pm$0.28} & 3.39\scriptsize{$\pm$0.34}  \\
				Spectral Clustering & \bf 91.12\scriptsize{$\pm$0.12} & \bf3.33 \scriptsize{$\pm$0.18} \\
				\bottomrule
		\end{tabular}
	}
	\end{center}
	\caption{Results under different clustering algorithms used for generating multi-centroid prototypes.}
	\label{table:clustering}
\end{table}

\noindent\textbf{Effectiveness of the query function}. In the design of our nearest neighbor query function, we surmise that data samples tend to locate near to their corresponding mass centers giving an appropriate embedding function, so it is important to verify that, given $r$ nearest neighbors, whether or not the target prompt falls in the candidate group. To this end, we demonstrate the top-$r$ accuracy 
under 5-centroid prototypes in Fig.~\ref{fig:top-r}. We can see that the top-$r$ accuracy increases monotonically according to the value of $r$ and $r=3$ works fairly well. Hence, a query vector in combination with key prototypes and a reasonable vicinity range can be effectively leveraged to retrieve the candidate task-specific prompts.

\section{Detailed Visualizations and Analysis}\label{appendix:visual} 
\begin{figure*}[t]
\begin{center}
	\includegraphics[width=0.40\textwidth]{figures/train_proto_train.jpg}
    \hfill
	\includegraphics[width=0.40\textwidth]{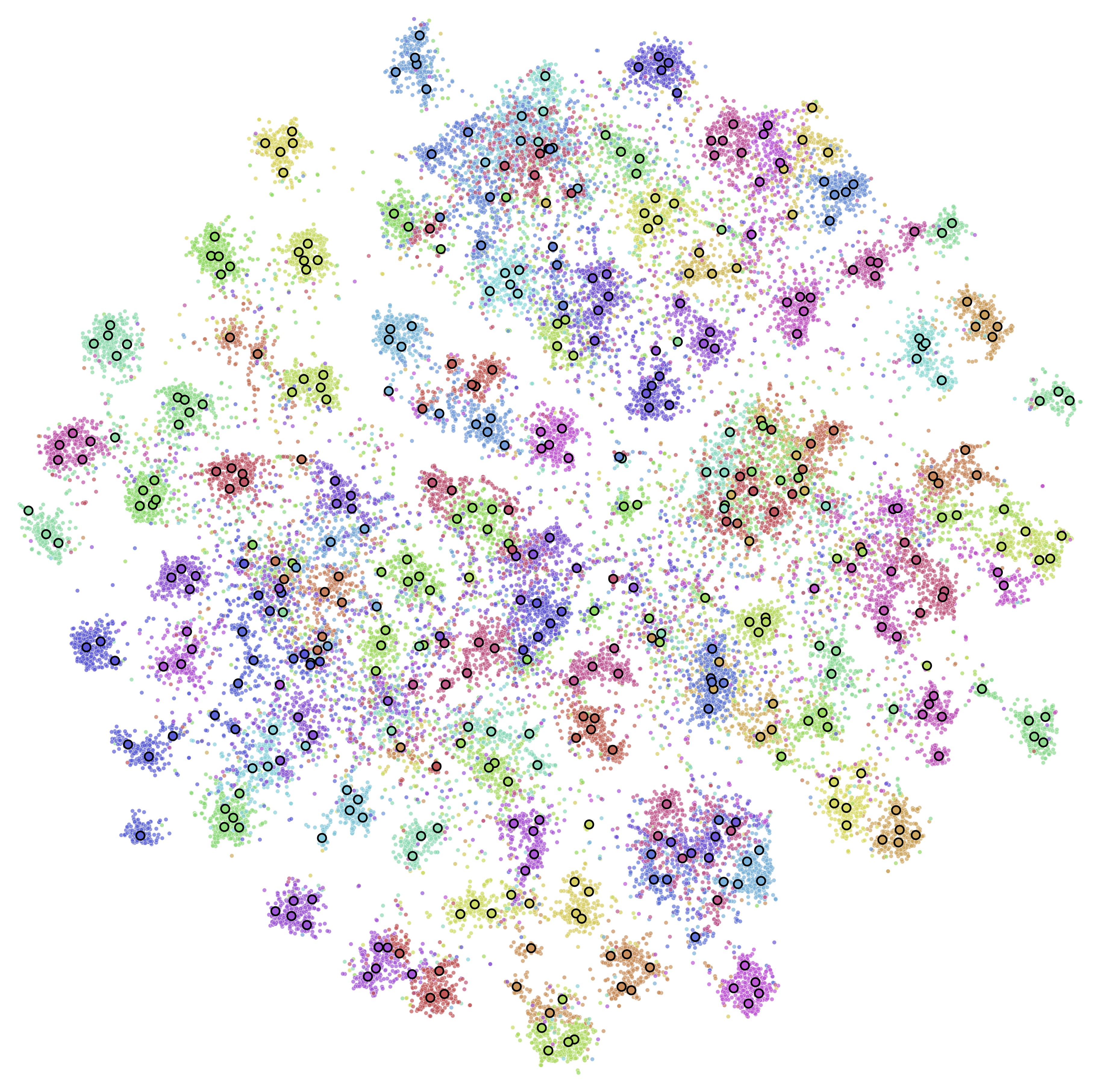}
	\hfill
	\includegraphics[width=0.40\textwidth]{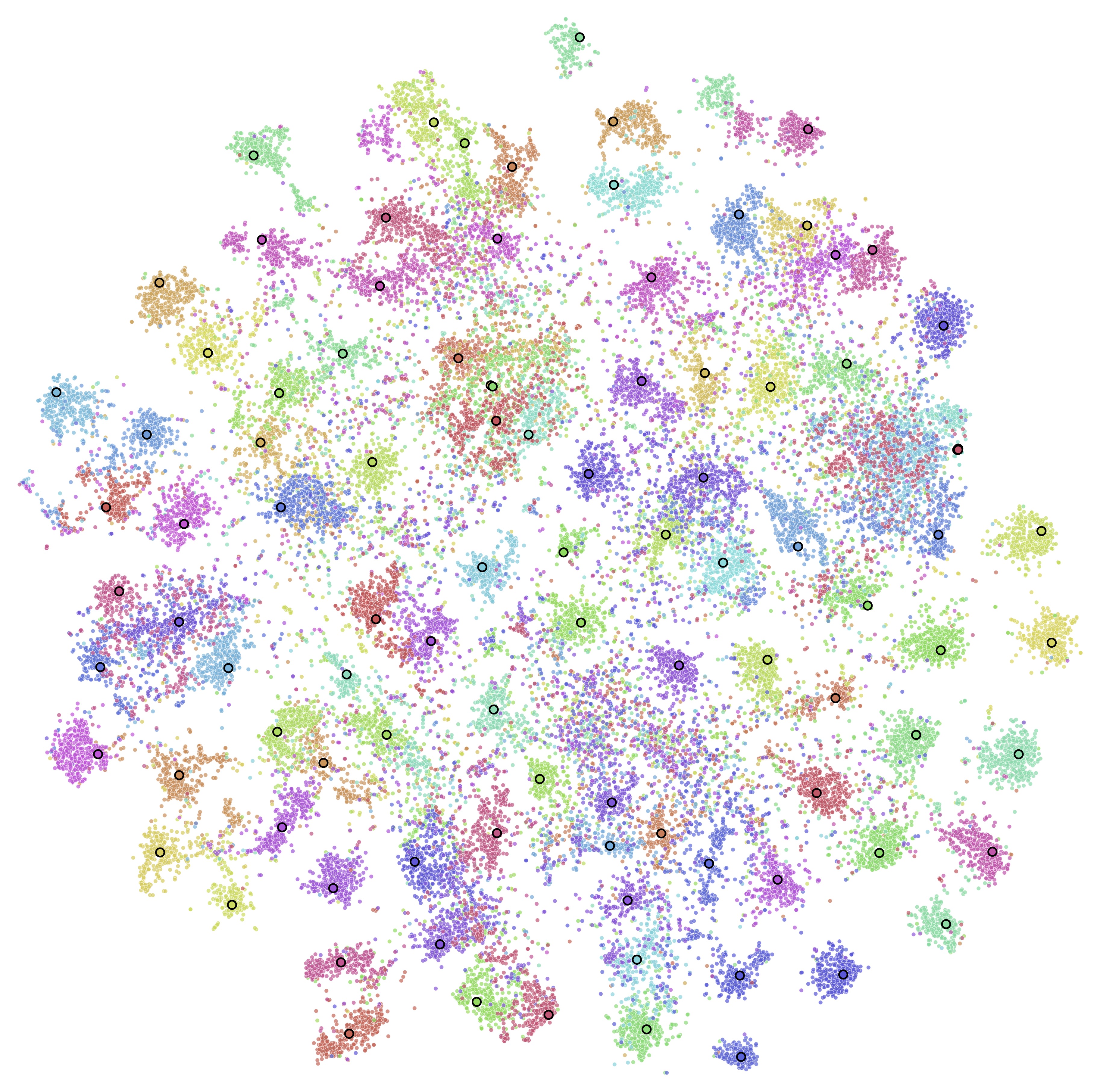}
	\hfill
	\includegraphics[width=0.40\textwidth]{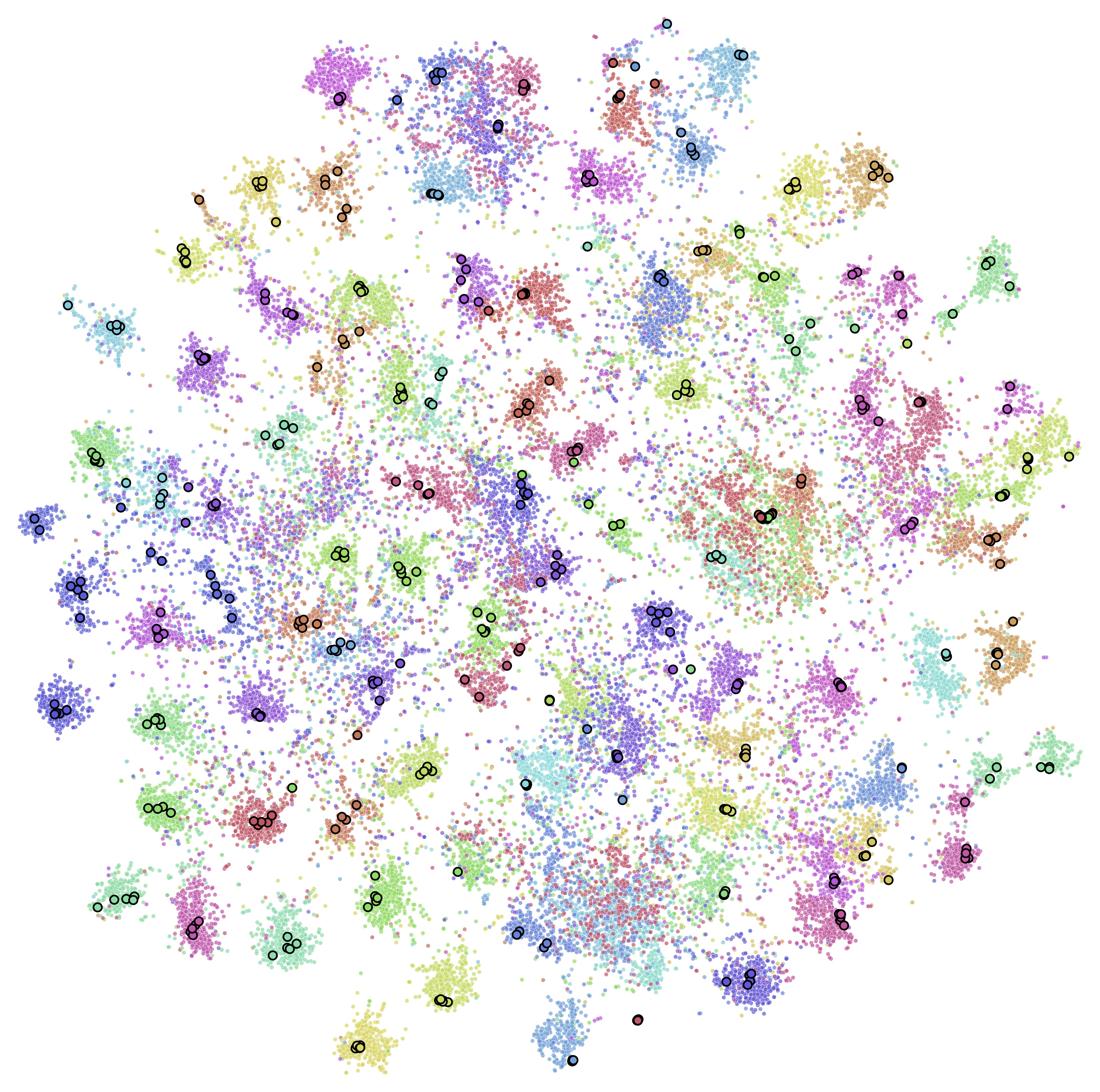}
	\hfill
	\includegraphics[width=0.40\textwidth]{figures/train_pt_proto_train_pt.jpg}
	\hfill
	\includegraphics[width=0.40\textwidth]{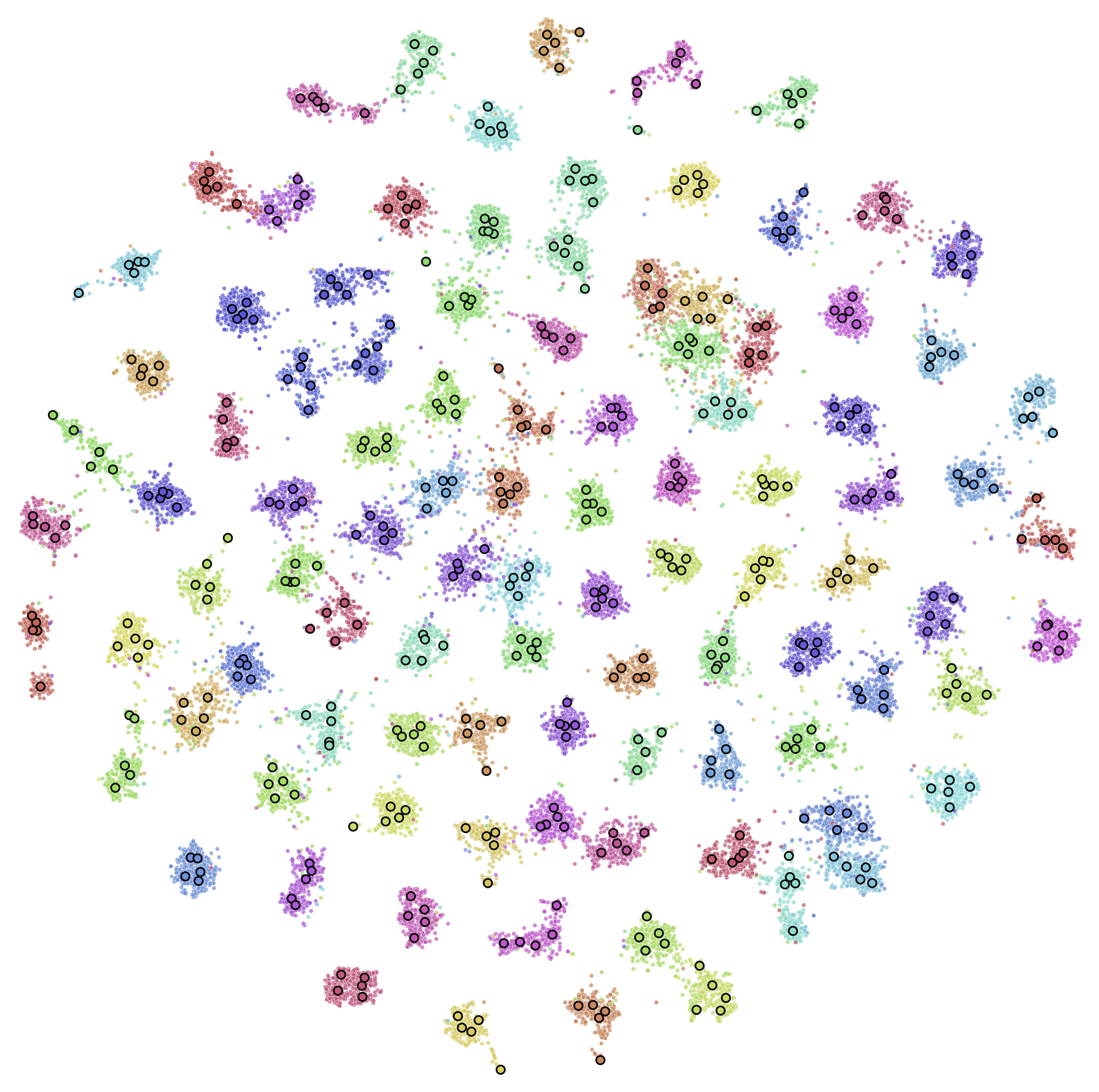}
\end{center}
\caption{Visualizations for train data in CIFAR-100. \textbf{First row}: Original data embeddings with class mean (left) and multi-centriod (right) \textit{key} prototypes. \textbf{Second row}: Original data embeddings with class mean (left) and multi-centriod (right) 
\textit{value} prototypes. \textbf{Third row}: Data embeddings after adding task-specific prompts with class mean (left) and multi-centriod (right) \textit{value} prototypes.}
\label{fig:train_visual}
\end{figure*}

Fig.~\ref{fig:train_visual} displays training samples from CIFAR-100 in the latent space. The first row shows original data samples with their corresponding class means and multi-centroid key prototypes. As shown in the figure, both class means and multi-centroid key prototypes effectively characterize the distribution of each class. In the second row, when replacing key prototypes with their corresponding value prototypes, there is a clear mismatch between the class distributions and their value prototypes. This observation manifests clear distribution shifts in the latent space after adding task-specific prompts, justifying the necessity of decoupling prototypes into the key prototypes and value prototypes. The third row exhibits value prototypes and embeddings of data samples after adding task-specific prompts. We can find that both class means and multi-centroid value prototypes fit the learned class distributions well. Nevertheless, multi-centroid value prototypes can better capture outliers, thus being more representative.

\begin{figure*}[t]
\begin{center}
	\includegraphics[width=0.40\textwidth]{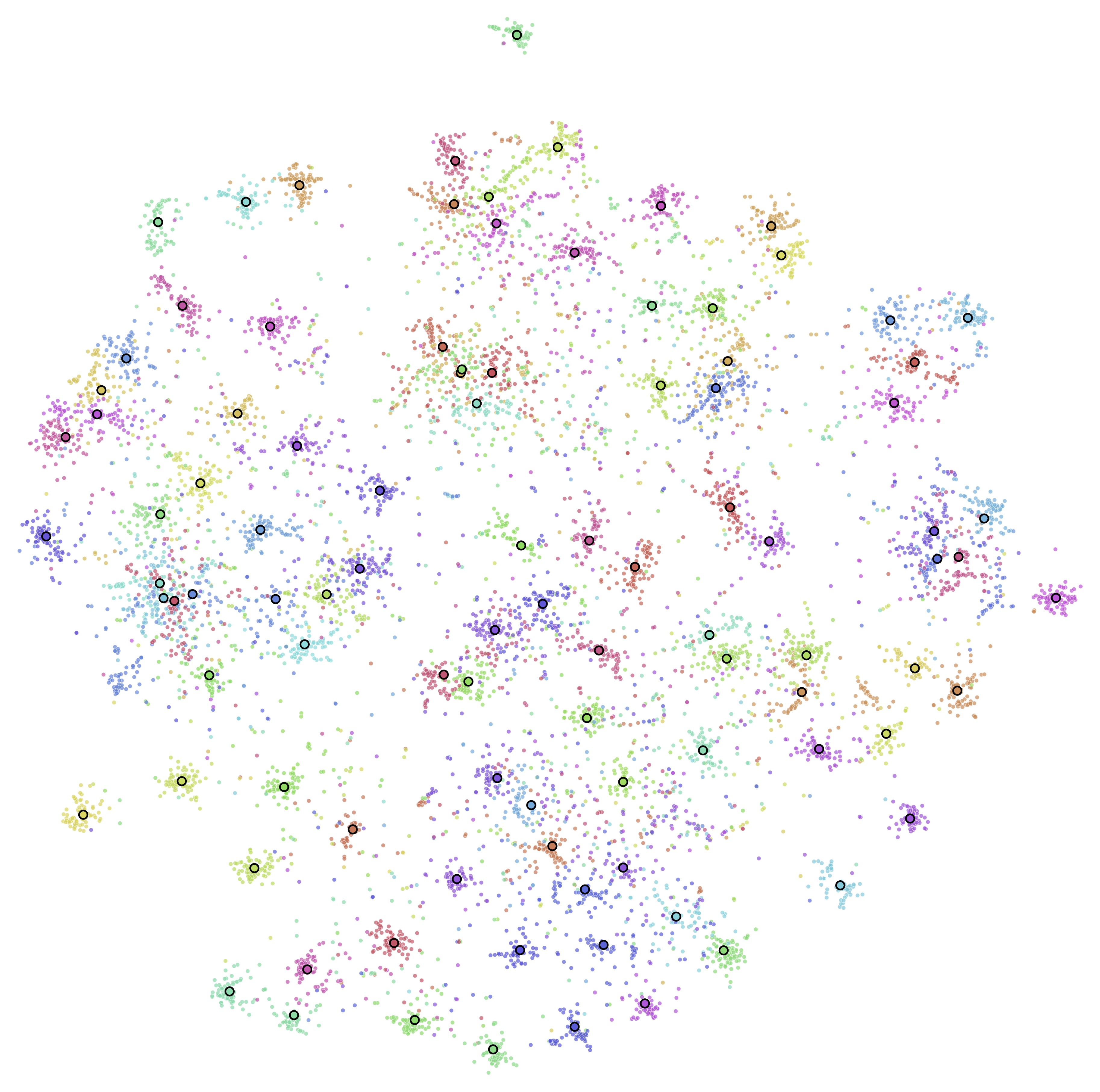}
	\hfill
	\includegraphics[width=0.40\textwidth]{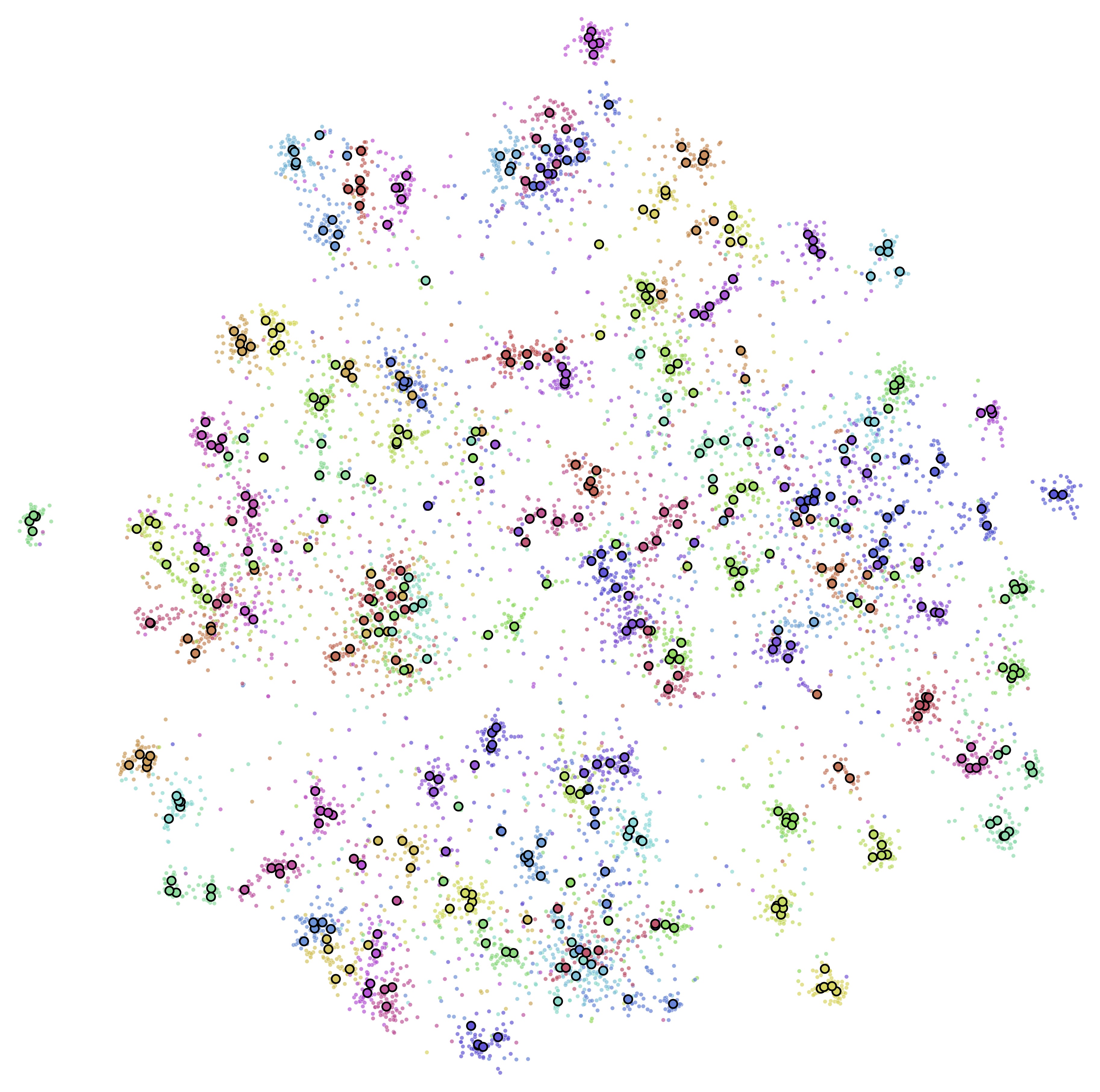}
	\hfill
	\includegraphics[width=0.40\textwidth]{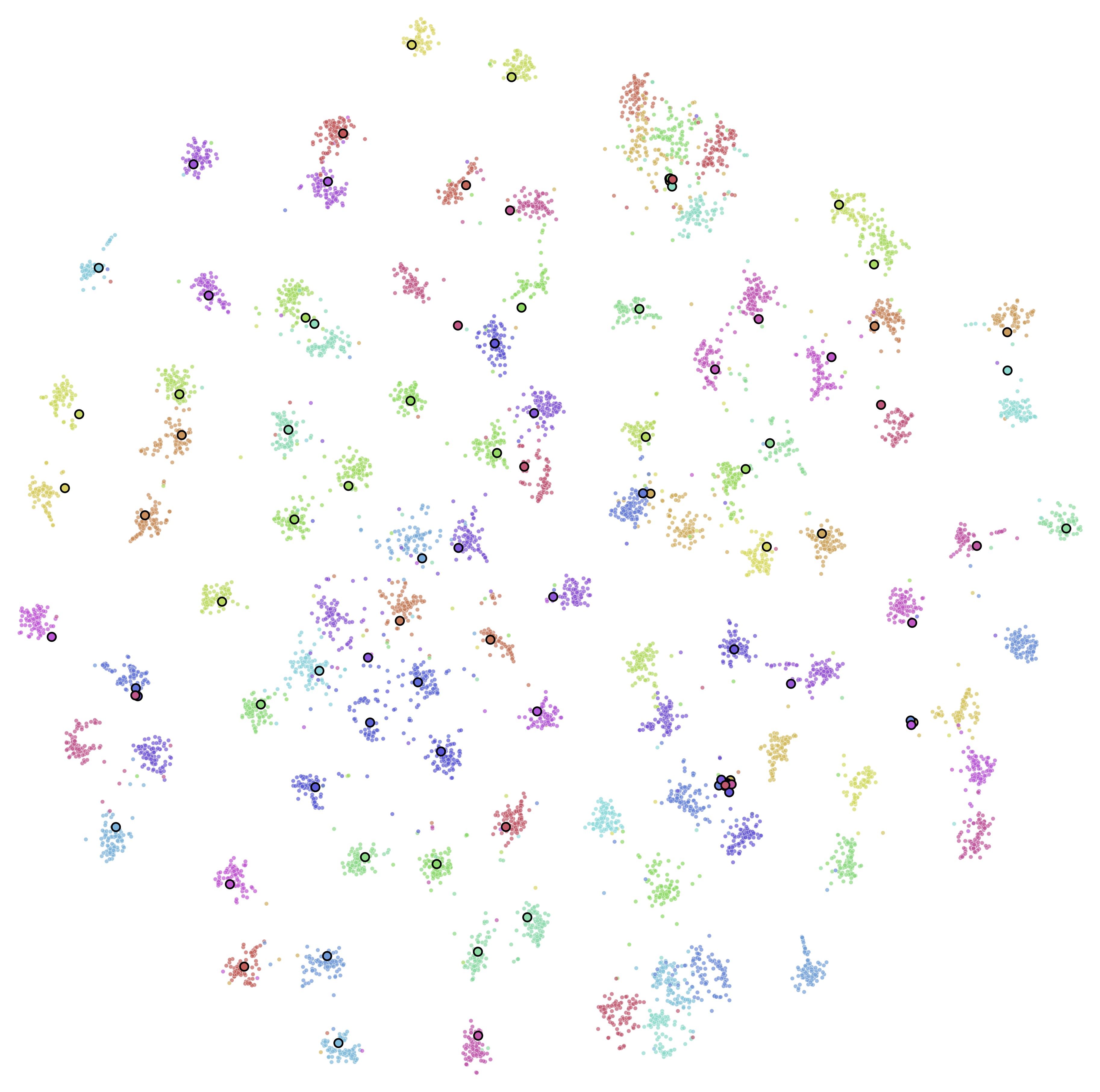}
	\hfill
	\includegraphics[width=0.40\textwidth]{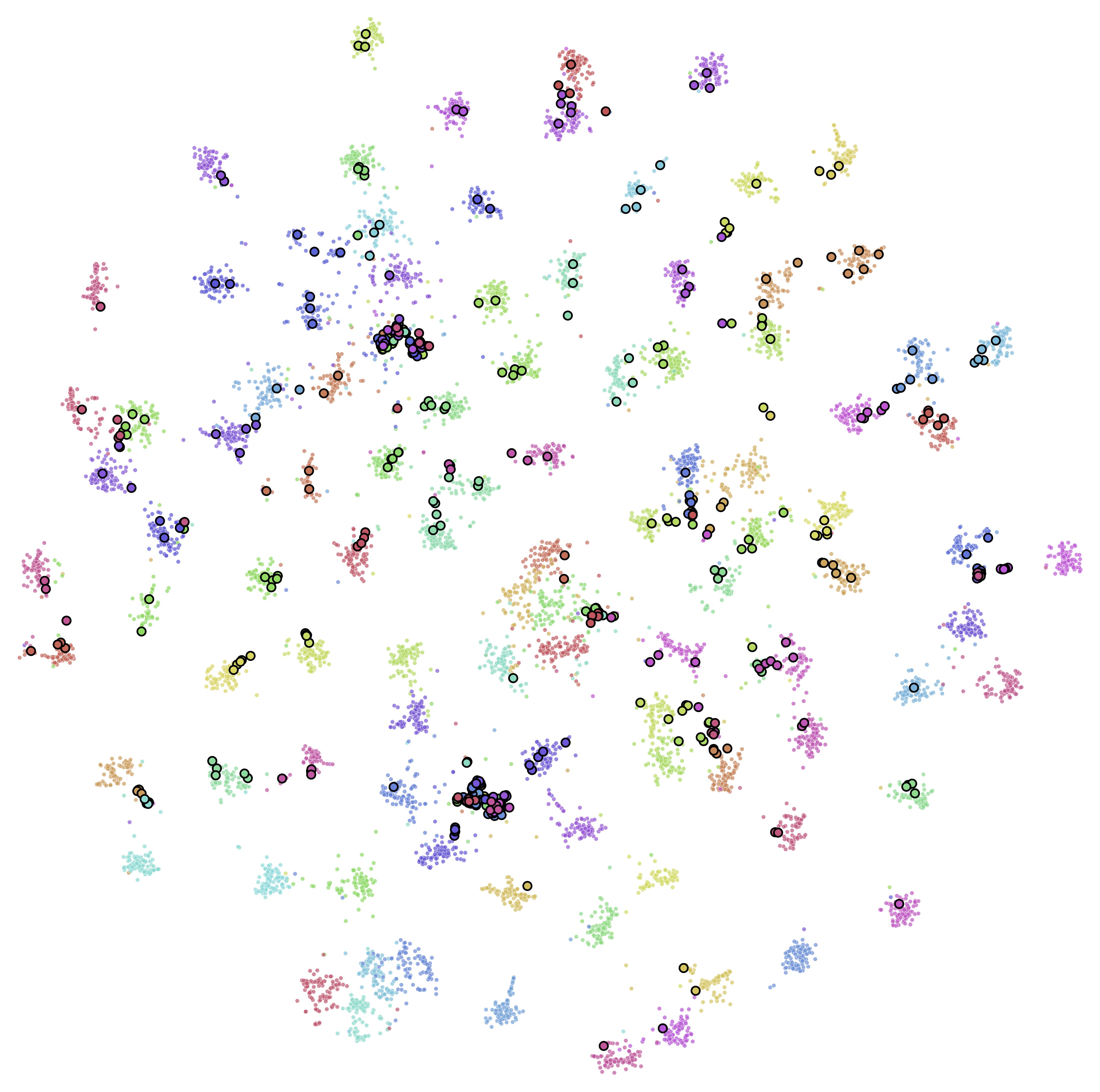}
	\hfill
	\includegraphics[width=0.40\textwidth]{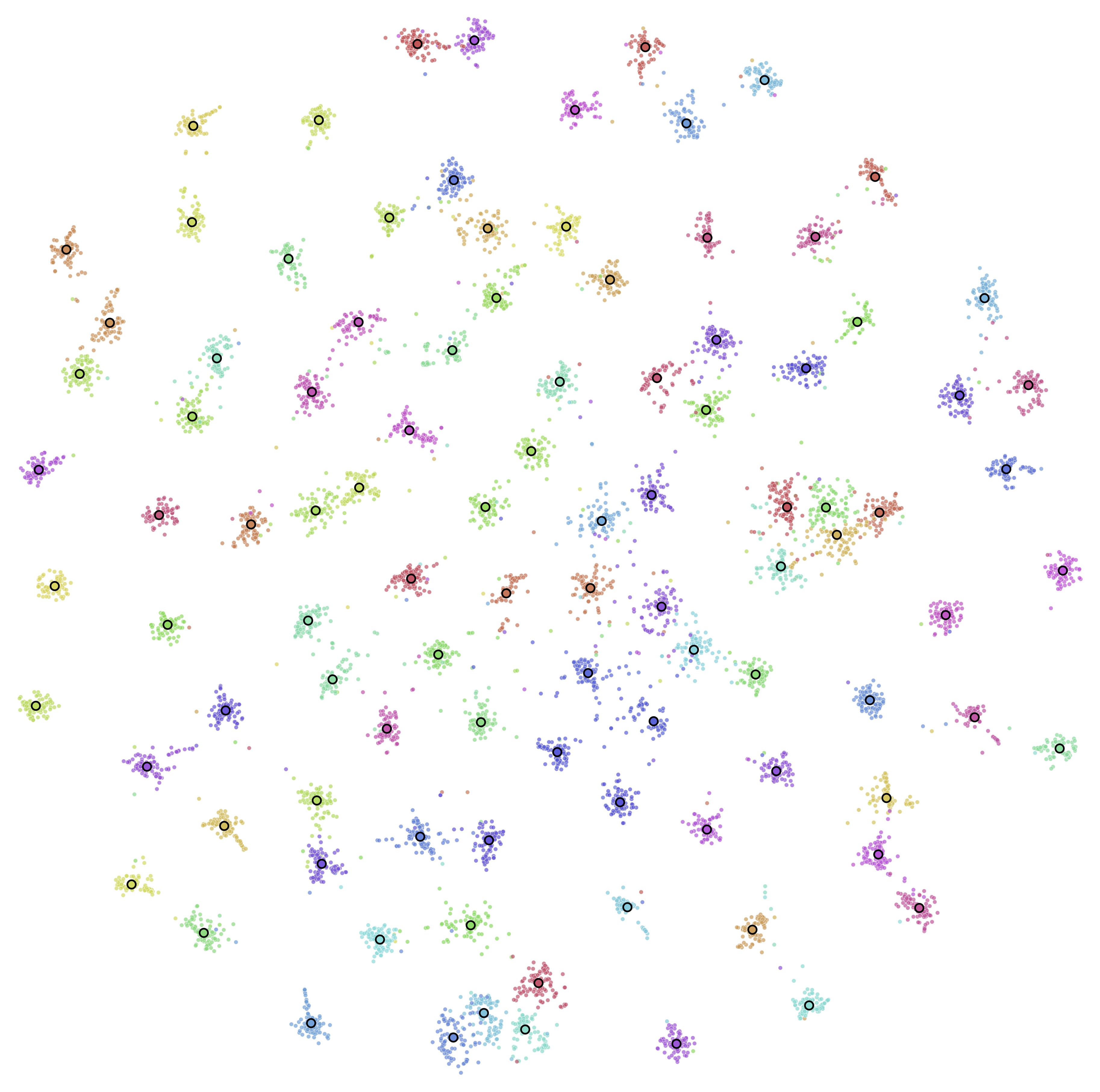}
	\hfill
	\includegraphics[width=0.40\textwidth]{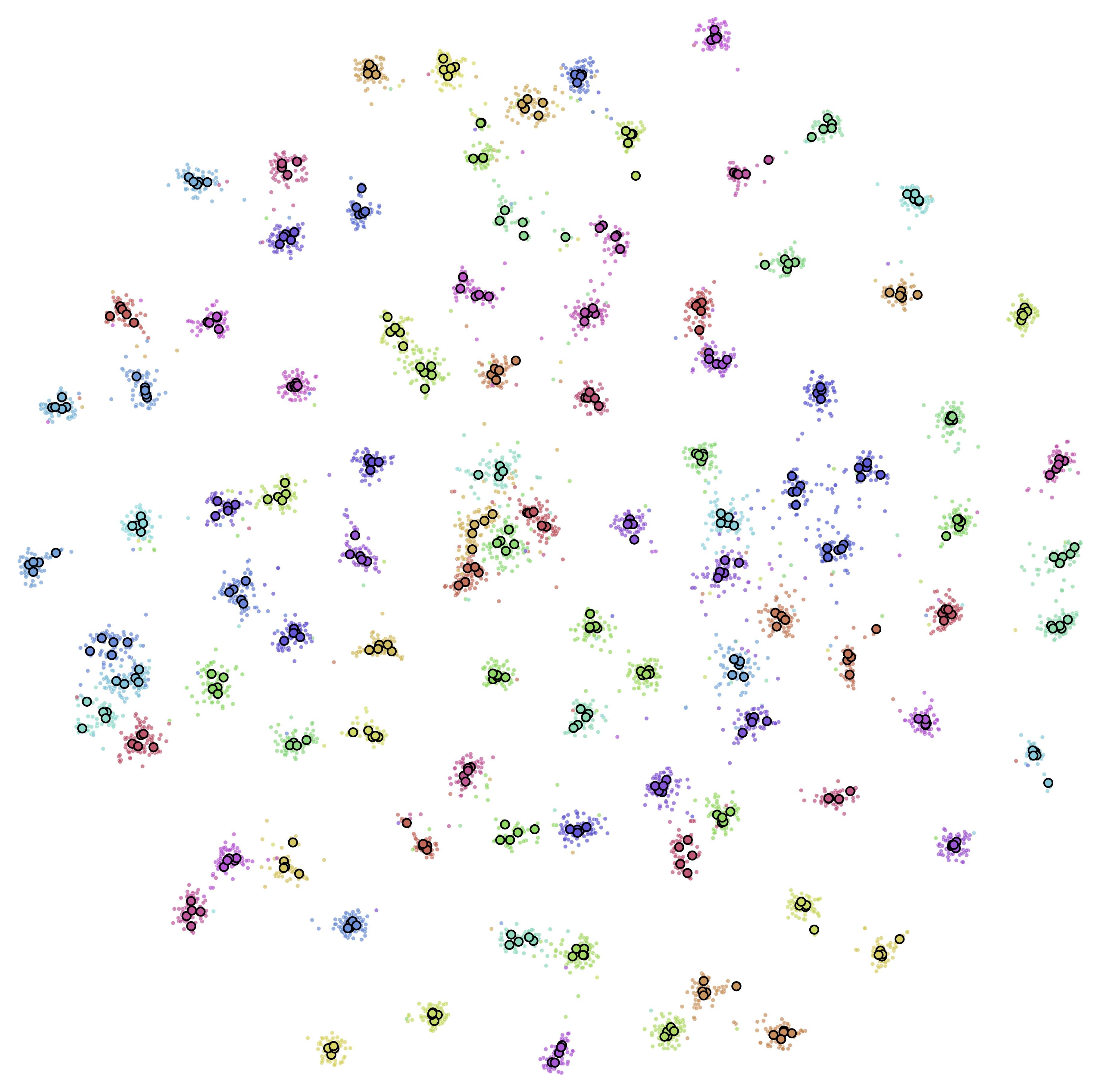}
\end{center}
\caption{Visualizations for test data in CIFAR-100. \textbf{First row}: Original data embeddings with class mean (left) and multi-centriod (right) \textit{key} prototypes. \textbf{Second row}: Original data embeddings with class mean (left) and multi-centriod (right) 
\textit{value} prototypes. \textbf{Third row}: Data embeddings after adding task-specific prompts with class mean (left) and multi-centriod (right) \textit{value} prototypes.}
\label{fig:test_visual}
\end{figure*}

In Fig.~\ref{fig:train_visual}, we show the efficacy of both key and value prototypes for representing train data. Here, in Fig.~\ref{fig:test_visual}, we exhibit their performances on test data. As shown in the first row, key prototypes work fairly well in representing the original test data embeddings and thus can be safely leveraged in the coarse retrieval process. The second row further validates the necessity of decoupling key and value prototypes from the perspective of test data. As shown in the second row, there are a few classes whose key prototypes can still effectively characterize their corresponding data distributions after task-specific prompts were added, suggesting minor distribution shifts. However, most classes fail to reuse key prototypes after adding task-specific prompts. In contrast, as shown the third row, value prototypes can always be reliably used as classifiers for predictions.

\end{document}

%% file: math_commands.tex
\usepackage{amsmath,amsfonts,bm}

\def\eqref#1{equation~\ref{#1}}

\def\1{\bm{1}}

\def\vmu{{\bm{\mu}}}

\def\vc{{\bm{c}}}

\def\ve{{\bm{e}}}

\def\vp{{\bm{p}}}
\def\vq{{\bm{q}}}

\def\vu{{\bm{u}}}

\def\vx{{\bm{x}}}

\def\vz{{\bm{z}}}

\def\mS{{\bm{S}}}

\DeclareMathAlphabet{\mathsfit}{\encodingdefault}{\sfdefault}{m}{sl}
\SetMathAlphabet{\mathsfit}{bold}{\encodingdefault}{\sfdefault}{bx}{n}

\newcommand{\R}{\mathbb{R}}

\DeclareMathOperator*{\argmin}{arg\,min}